\documentclass[11pt,letterpaper]{article}

\usepackage{cogsys}

\usepackage{times}
\usepackage[pdftex]{graphicx} 

\usepackage{natbib}
\cogsysheading{}{2018}{}{}{}

\ShortHeadings{{\sffamily\footnotesize Human-Centred Explainability of Visuo-Spatial Symmetry}}
              {{\sffamily\footnotesize J.\ Suchan, M.\ Bhatt, S.\ Varadarajan, A.\ Amirshahi, S.\ Yu}}
              
\usepackage{xcolor}
\usepackage{soul}

\usepackage[small]{caption}

\usepackage{amsmath}
\usepackage{amssymb}
\usepackage{subcaption}
\usepackage{float}
\usepackage[ruled,vlined,linesnumbered]{algorithm2e}

\usepackage{graphicx}
\usepackage[export]{adjustbox}

\usepackage{array}
\usepackage{booktabs}
\usepackage{multirow}
\usepackage{pifont}

\newcolumntype{f}{>{$}l<{$}}
\newcolumntype{n}{l}
\newcolumntype{N}{>{\scriptsize}l}
\newcolumntype{v}[1]{>{\raggedright\hspace{0pt}}p{#1}}
\newcolumntype{V}[1]{>{\scriptsize\raggedright\hspace{0pt}}p{#1}}
\newcolumntype{C}[1]{>{\scriptsize\centering\hspace{0pt}}p{#1}}
\newcolumntype{L}[1]{>{\scriptsize\hspace{0pt}}p{#1}}
\newcolumntype{B}[1]{>{\boldmath\DC@{.}{,}{#1}}l<{\DC@end}}
\newcolumntype{d}[1]{>{\DC@{.}{,}{#1}}l<{\DC@end}}
\newcolumntype{R}[1]{%
>{\begin{turn}{90}\begin{minipage}{#1}\scriptsize\raggedright\hspace{0pt}}l%
<{\end{minipage}\end{turn}}%
}
\newcolumntype{x}{>{\scriptsize\raggedright\hspace{0pt}}X}

\newcommand{\predThF}[1]{{\operatorname{\mathsf{#1}}}}

\newcommand{\href}[2]{\url{#1 }}

\usepackage[pagebackref=true,breaklinks=true,letterpaper=true,colorlinks,bookmarks=false]{hyperref}

\hypersetup{
    pdftitle={Semantic Analysis of (Reflectional) Visual Symmetry: A Human-Centred Computational Model for Declarative Explainability}, 
    pdfauthor={Jakob Suchan, Mehul Bhatt},     	
    pdfsubject={cognitive vision, deep semantics, visual art, digital media}, 
    pdfkeywords={vision} {AI} {explainability}, 					
    unicode=false,          									
    pdftoolbar=false,        									
    pdfmenubar=true,        								
    pdffitwindow=false,     									
    pdfstartview={FitH},    								
    pdfnewwindow=true,      								
    bookmarks=true,         								
    colorlinks=true,       									
    linkcolor=red!90!black,          							
    citecolor=blue!90!black,        							
    filecolor=blue,      									
    urlcolor=blue!90!white          								
}

\begin{document} 

\title{{\sffamily\large\color{blue!50!black}\uppercase{Semantic Analysis of (Reflectional) Visual Symmetry}}\\[5pt]{\color{blue!45!black}\small\sffamily A Human-Centred Computational Model for Declarative Explainability}}

{
\footnotesize
\sffamily

\author{\scriptsize\sffamily Jakob Suchan}{\url{info@codesign-lab.org}}

\author{\scriptsize\sffamily Mehul Bhatt}{}

\author{\scriptsize\sffamily Srikrishna Varadarajan}{}

\address{\scriptsize\sffamily CoDesign Lab EU $~$/$~$ Cognitive Vision\\\url{www.codesign-lab.org} $~$/$~$ \url{www.cognitive-vision.org}\\[4pt]\"{O}rebro University, Sweden $~$---$~$ University of Bremen, Germany

}

\author{\scriptsize\sffamily Seyed Ali Amirshahi}{}
\address{Norwegian Colour and Visual Computing Laboratory\\Norwegian University of Science and Technology, Norway}

\author{\scriptsize\sffamily Stella Yu}{}
\address{ICSI Vision Group, UC Berkeley, United States}

}

\vskip 0.5in
 
\begin{abstract}We present a computational model for the semantic interpretation of symmetry in naturalistic scenes. Key features include a human-centred representation, and a declarative, explainable interpretation model supporting deep semantic question-answering founded on an integration of methods in knowledge representation and deep learning based computer vision. In the backdrop of the visual arts, we showcase the framework's capability to generate human-centred, queryable, relational structures, also evaluating the framework with an empirical study on the human perception of visual symmetry. Our framework represents and is driven by the application of foundational, integrated Vision and Knowledge Representation and Reasoning methods for applications in the arts, and the psychological and social sciences.
\end{abstract}

\section{Introduction}
\begin{figure}
	\centering
	
	\begin{subfigure}[b]{0.28\linewidth}
    		\includegraphics[height=6.57cm]{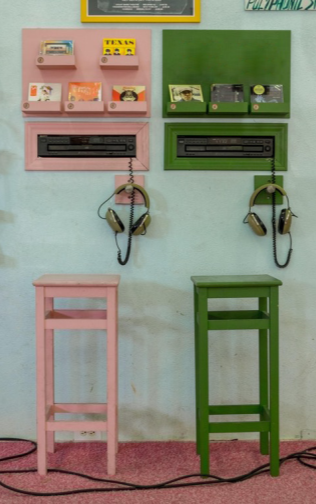}
		\caption{}
		\label{fig:headphones}
    	\end{subfigure}
    	\begin{subfigure}[b]{0.70\linewidth}
        		\includegraphics[height=3.05cm]{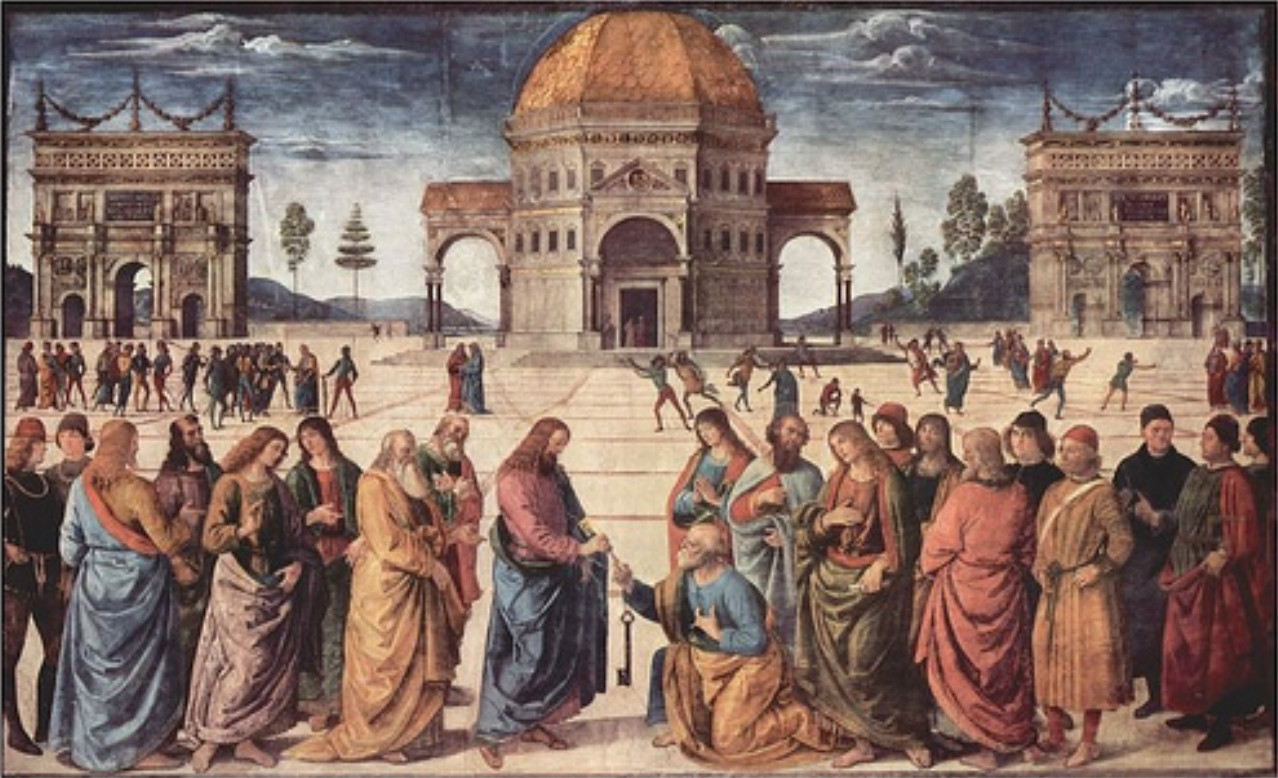}
		\includegraphics[height=3.05cm]{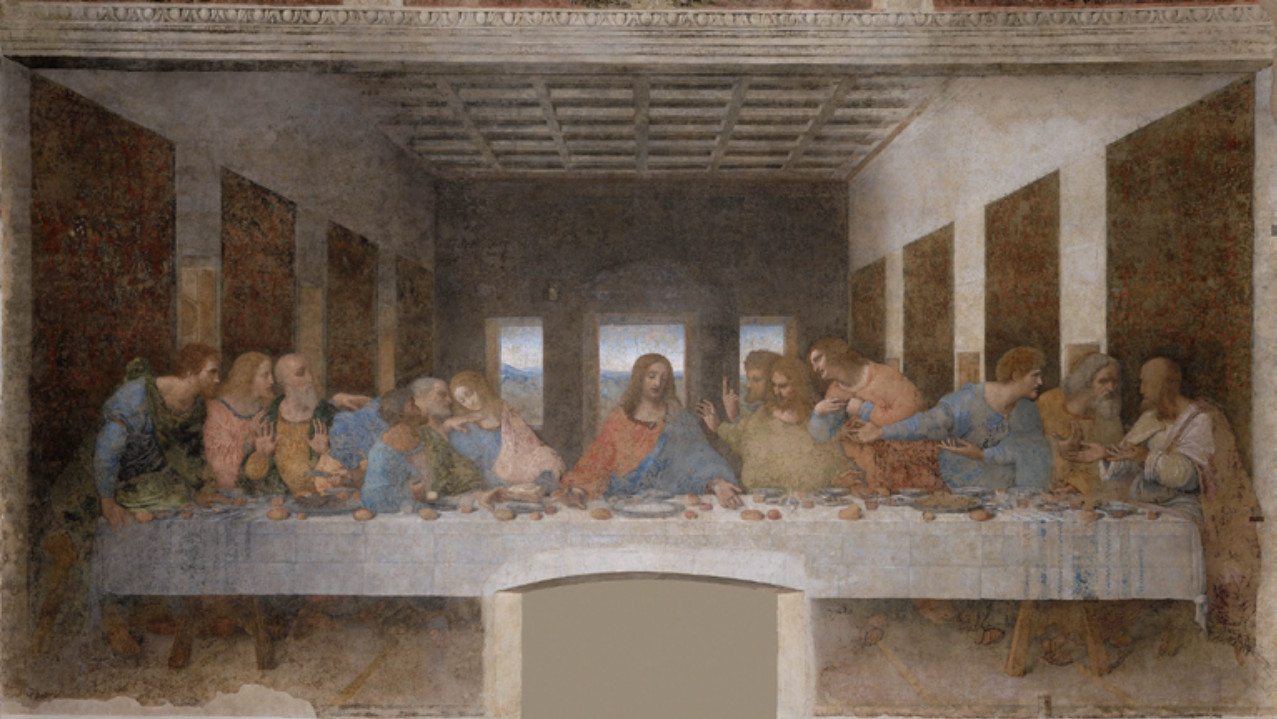}\\[5pt]
    		\includegraphics[height=3.35cm]{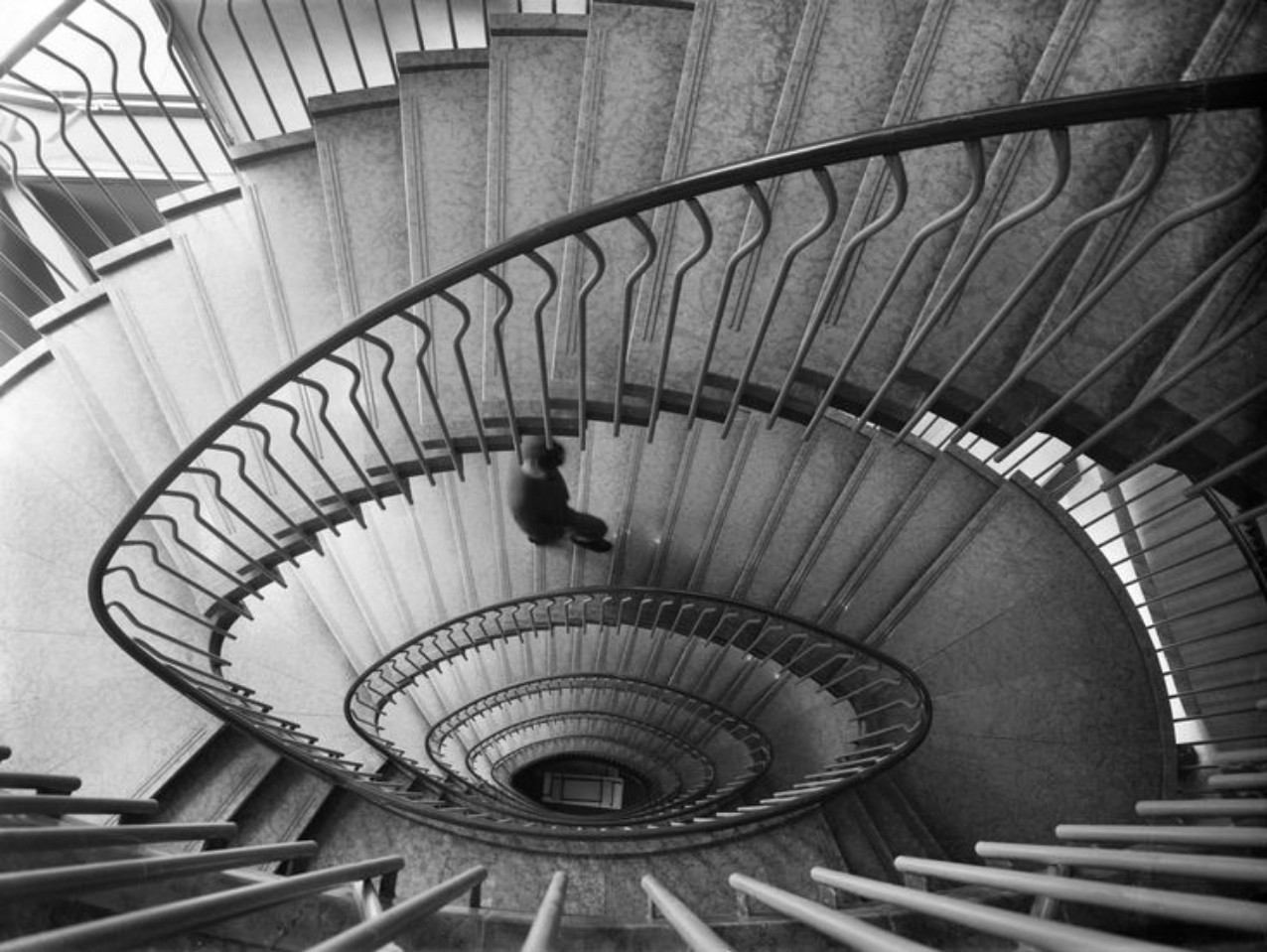}
		\includegraphics[height=3.35cm]{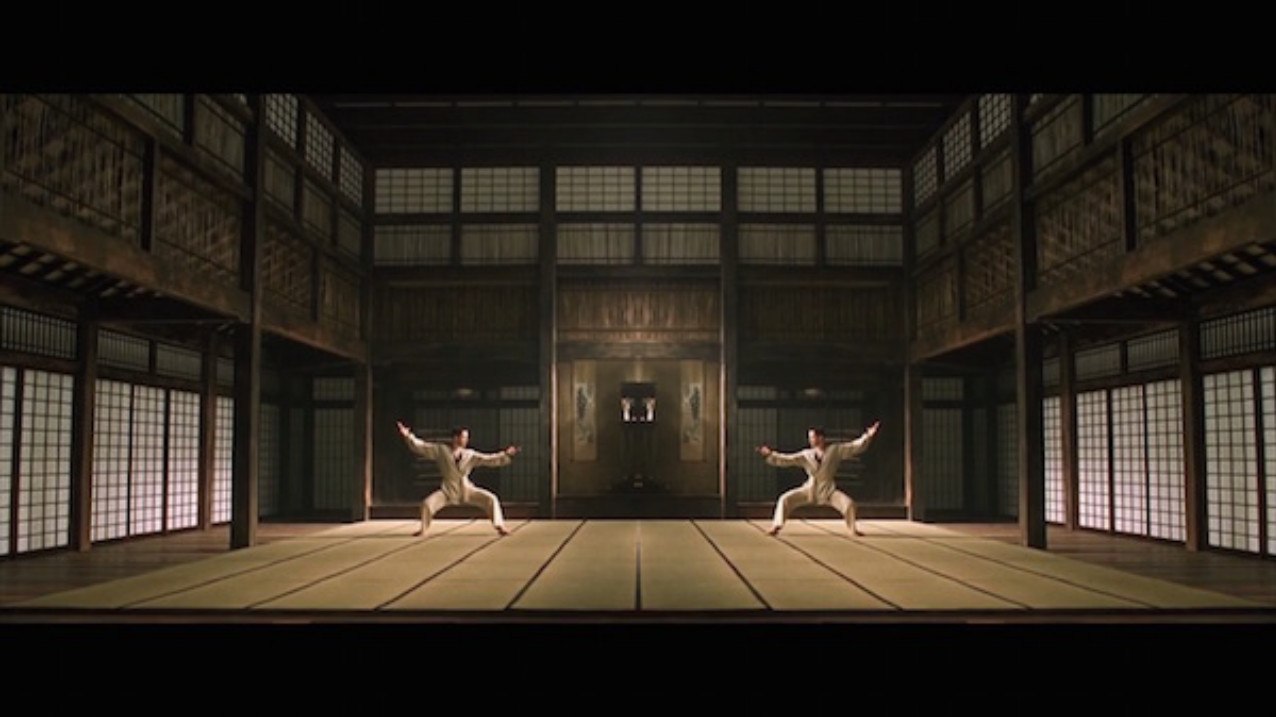}
		\caption{}
 	\end{subfigure}
 
	\caption{{ The perception of symmetry. (a) Symmetry perception influenced by visual features, conceptual categories, semantic layering, and nuances of individual differences in perception, and (b) examples for symmetry in visual arts: \emph{``Delivery of the Keys''} (ca.1481) by Perugino, \emph{``The Last Supper''} (1495-98) by Leonardo Da Vinci, \emph{``View of the grand staircase at La Rinascente in Rome, designed by Franco Albini and Franca Helg''} (1962) by Giorgio Casali, and \emph{``The Matrix''} (1999) by The Wachowski Brothers.}} 	
	\label{fig:sample-sym-first-page}
\end{figure}

Visual symmetry as an aesthetic and stylistic device has been employed by artists across a spectrum of creative endeavours concerned with visual imagery in some form, e.g., painting, photography, architecture, film and media design. Symmetry in (visual) art and beyond is often linked with elegance, beauty, and is associated with attributes such as being well-proportioned and well-balanced \citep{weyl1952symmetry}. Closer to the ``visual imagery'' and ``aesthetics'' centred scope of this paper, symmetry has been employed by visual artists going back to the masters {\small\sffamily Giorgione, Titian, Raphael, da Vinci}, and continuing till the modern times with {\small\sffamily Dali} and other contemporary artists (Figure \ref{fig:sample-sym-first-page}).

\medskip

\textbf{Visual Symmetry: Perception and Semantic Interpretation}\quad There exist at least four closely related points of view pertaining to symmetry, namely, the physical, mathematical, pyschological, and aesthetical points of view \citep{Symmetry-visart-MOLNAR}. As \citet{Symmetry-visart-MOLNAR} articulate:

\begin{quote}
{``But perceptual symmetry is not always identical to the symmetry defined by the mathematicians. A symmetrical picture is not necessarily symmetrical in the mathematical sense...Since the aesthetical point of view is strictly linked to the perceptive system, in examining the problems of aesthetics we find ourselves dealing with two distinct groups of problems: (1) the problem of the perception of symmetry; (2) the aesthetical effect of the perception of a symmetrical pattern.''}
\end{quote}

Indeed, the high-level semantic interpretation of symmetry in naturalistic visual stimuli by humans is a multi-layered perceptual phenomena operating at several interconnected cognitive levels involving, e.g., spatial organisation, visual features, semantic layers, individual differences (Section \ref{subsec:multilevel}; and Figure \ref{fig:headphones}). Consider the select examples from movie scenes in Figure \ref{fig:symmetry_analysis}:

\begin{figure*}[t]
\centering
\scriptsize
\begin{subfigure}[b]{\linewidth}
	\centering
        \subcaptionbox{\label{subfig:kubrick_sym}}{\includegraphics[height=2.29cm]{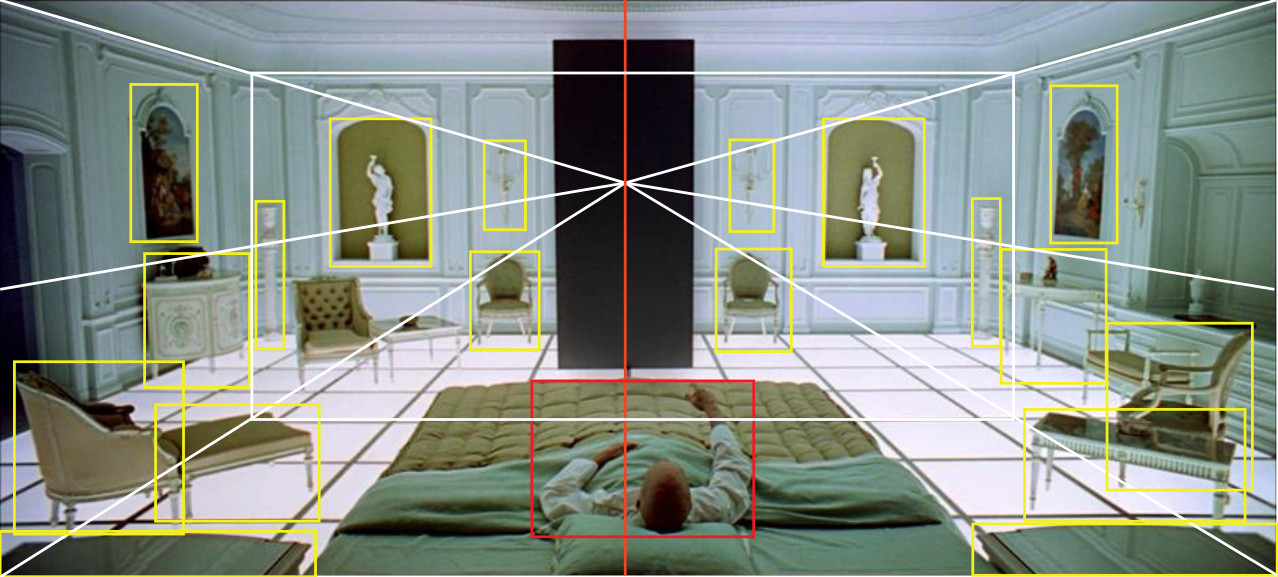}}\quad
    \subcaptionbox{\label{subfig:anderson_sym}}{\includegraphics[height=2.29cm]{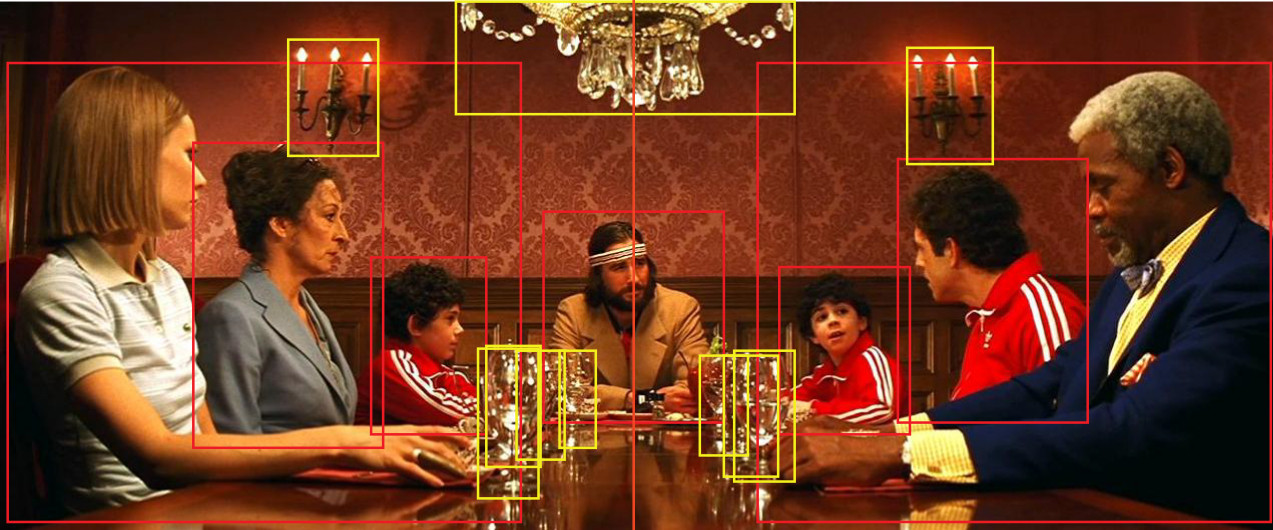}}\quad
     \subcaptionbox{\label{subfig:coen_sym}}{\includegraphics[height=2.29cm]{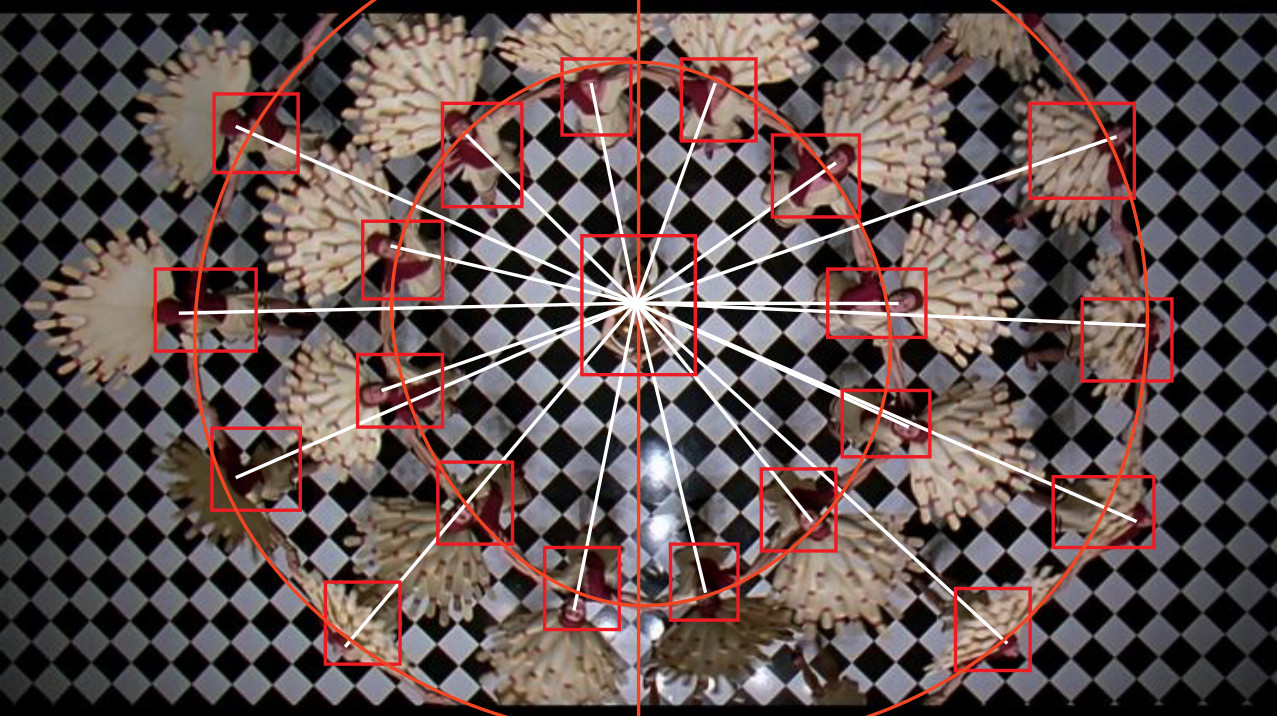}}
\end{subfigure}
\caption{{ Symmetrical structure in visual arts. Select scenes from films: (a) \emph{``2001: A Space Odyssey''} (1968) by Stanley Kubrick, (b) \emph{``The Royal Tenenbaums''} (2001) by Wes Anderson, and (c) \emph{``The Big Lebowski''} (1998) by Joel and Ethan Coen}.}
\label{fig:symmetry_analysis}
\end{figure*}%

\begin{itemize}

\item in the shot from {\sffamily\small``2001: A Space Odyssey''} (Figure \ref{subfig:kubrick_sym}) a centre-perspective is being applied for staging the scene. The symmetry here is obtained by this, as well as by the layout of the room, the placement of the furniture, and the decoration of the room. In particular, the black obelisk in the centre of the frame is emphasising the centre-perspective regularly used by Kubrick, with the bed (and person) being positioned directly on the central axis.

\item {\sffamily\small Wes Anderson} is staging his shot from {\sffamily\small ``The Royal Tenenbaums''} (Figure \ref{subfig:anderson_sym}) around a central point, but unlike {\small\sffamily Kubrick's} shot, {\sffamily\small Anderson} focuses on the people involved in it. Even though the visual appearance of the characters differs a lot, the spatial arrangement and the semantic similarity of the objects in the shot creates symmetry. Furthermore, the gazing direction of the characters, i.e., people on the right facing left and people on the left facing right, adds to the symmetrical appearance of the shot.

\item In {\sffamily\footnotesize``The Big Lebowski''} (Figure \ref{subfig:coen_sym}), Joel and Ethan Coen use symmetry to highlight the surreal character of a dream sequence; the shot in Figure \ref{subfig:coen_sym} uses radial symmetry composed of a group of dancers, shot from above, moving around the centre of the frame in a circular motion. This is characterised by moving entities along a circular path and centre-point, and the perceptual similarity in the appearance of the dancers. 

\end{itemize}

The development of computational cognitive models focussing on a human-centred --\emph{semantic, explainable}-- interpretation of visuo-spatial symmetry presents a formidable research challenge demanding an interdisciplinary ---mixed-methods--- approach at the interface of cognitive science, vision \& AI, and visual perception focussed human-behavioural research. Broadly, our research is driven by addressing this interdisciplinarity, with an emphasis on developing integrated KR-and-vision foundations for applications in the psychological and social sciences, e.g., archival, automatic annotation and pre-processing for qualitative analysis, studies in visual perception.

\medskip


\textbf{Key Contributions}\quad  
The core focus of the paper is to present a computational model with the capability to generate semantic, explainable interpretation models for the analysis of visuo-spatial symmetry. The explainability is founded on a domain-independent, mixed qualitative-quantitive representation of visuo-spatial relations based on which the symmetry is declaratively characterised. We also report on a qualitative evaluation with human-subjects, whereby human subjects rank their subjective perception of visual symmetry for a set stimuli using (qualitative) distinctions. The broader implications are two-fold: (1) the paper  demonstrates the integration of vision and semantics, i.e., knowledge representation and reasoning methods with low-level (deep learning based) visual processing methods; and (2). from an applied viewpoint, the developed methodology can serve as the technical backbone for assistive and analytical technologies for visual media studies, e.g., from the viewpoint of psychology, aesthetics, cultural heritage.



\begin{figure*}[t]
	\centering

    \includegraphics[width=\textwidth]{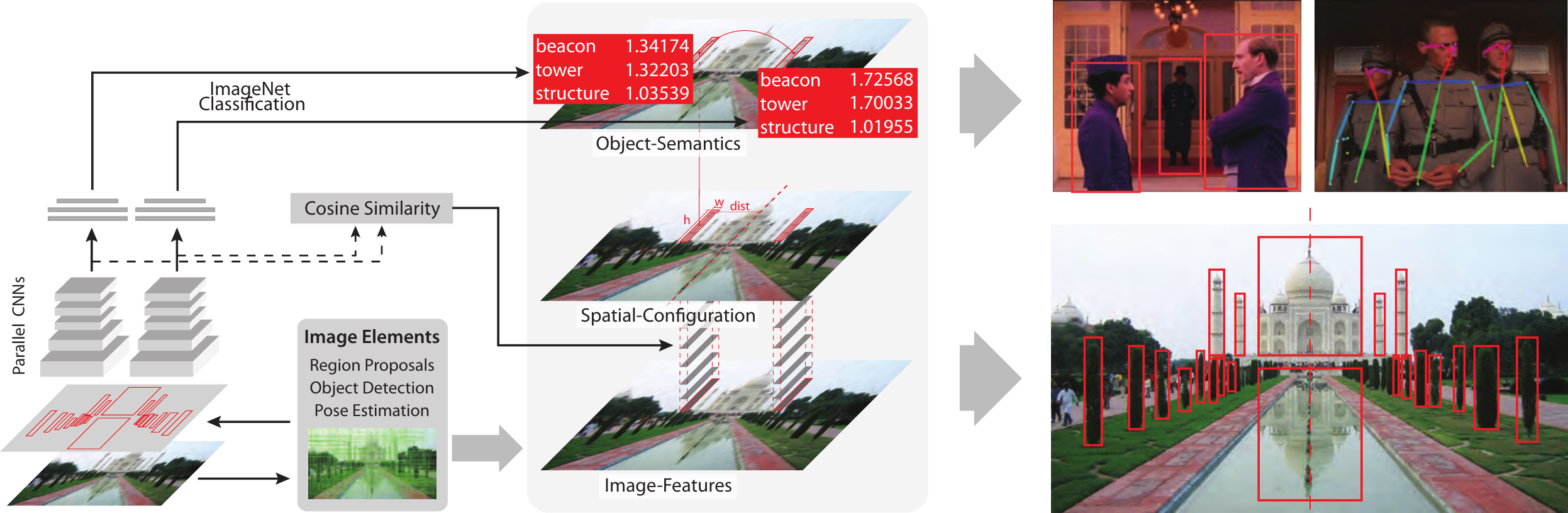}

	\caption{{ A computational model of multi-level semantic symmetry.}} 		
			
	\label{fig:computational_symmetry}
\end{figure*}

\section{The Semantics of Symmetry}\label{sec:symsemantic}

\noindent Symmetry in visual imagery denotes that an image is invariant to certain types of transformation of the image, e.g., reflectional symmetry is the case where the image does not change, when it is mirrored along a specific symmetry-axis.
Besides reflectional symmetry, there are various types of symmetry, including rotational symmetry, and translational symmetry. Perfect symmetry can be easily detected based on image level features, by comparing pixel in the image; however, in natural images, e.g., coming from the visual arts, perfect symmetry is a very rare case and mostly variations of symmetry are used as a stylistic device, with it being present only in some parts of the image.
To address this, we focus on developing a semantic model capable of interpreting symmetrical structures in images.

\subsection{A Multi-Level Semantic Characterisation}\label{subsec:multilevel}
From the viewpoint of perceptual and aesthetic considerations, key aspects for interpreting visual-spatial symmetry (in scope of the approach of this paper) include ({\small\color{black}S1--S4}; Figure \ref{fig:sample-sym-first-page}): 

\medskip

({\sffamily{\color{black}\footnotesize S1}}) {\bf \small Spatial organisation}:\quad High-level conceptual categories identifiable from geometric constructions by way of arbitrary shapes, relative orientation and placement, size of geometric entities, relative distance, and depth

\smallskip

({\sffamily{\color{black}\footnotesize S2}}) {\bf\small Visual features}:\quad Low-level visual features and artefacts emanating directly from color, texture, light, and shadow

\smallskip

({\sffamily{\color{black}\footnotesize S3}}) {\bf\small Semantic layers}:\quad Semantic-spatial layering and grouping based on natural scene characteristics involving, for instance,  establishing foreground-background, clustering based on conceptual similarity, relative distance, and perceived depth, and application of commonsense knowledge possibly not directly available in the stimulus

\smallskip

({\sffamily{\color{black}\footnotesize S4}}) {\bf\small Individual differences}:\quad Grounding of the visual features in the socio-cultural semiotic landscape of the perceiver (i.e., contextual and individualised nuances in perception and sensemaking).

\medskip

We develop a multi-level characterisation of symmetry aimed at analysing (reflectional) symmetry. Visual symmetry ---in this paper--- encompasses three layers ({\small\color{black}L1--L3}; Figure \ref{fig:computational_symmetry}): 

\medskip

\noindent{\sffamily\footnotesize{\bf\color{black} L1}}.$~$ \emph{Symmetrical (spatial) composition}:\quad Spatial arrangement of objects in the scene with respect to a structural representation of a wrt. position, size, orientation, etc.;

\smallskip

\noindent{\sffamily\footnotesize{\bf\color{black} L2}}.$~$ \emph{Perceptual similarity}: \quad Perceptual similarity of features in symmetrical image patches, based on the low-level feature based appearance of objects, e.g., colour, shape, patterns, etc.;

\smallskip

\noindent{\sffamily\footnotesize{\bf\color{black} L3}}.$~$ \emph{Semantic similarity}: \quad Similarity of semantic categories of the objects in symmetrical image patches, e.g., people, object types, and properties of these objects, such as peoples gazing direction, foreground / background etc.

\smallskip

The proposed characterisation serves as the foundation for analysing and interpreting symmetrical structures in the images; in particular it can be used to identify the elements of the image supporting the symmetrical structure, but also those parts of the image that are not in line with the symmetry, e.g., elements breaking the symmetry. This may be used for investigating the use of balance and in-balance in visual arts, and for analysing how this can be used to guide peoples attention in the context of visual saliency.

\subsection{{A Model of Reflectional Symmetry}}\label{sec:reflectionsym}
\noindent For the computational model presented in this paper (Figure \ref{fig:computational_symmetry}), we focus on reflectional symmetry in the composition of the image based on layers {\small\color{black}L1--L3} (Section \ref{subsec:multilevel}), i.e., we investigate image properties based on spatial configuration, low-level feature similarity, and semantic similarity. 
Towards this we extract image elements $\mathcal{E}_{1\cup2\cup3} = \{e_0, ..., e_n\}$ of the image: \quad

\begin{itemize}

\item[($\mathcal{E}_1$)] \emph{Image patches} are extracted using selective search as described in \cite{Uijlings2013}; resulting in structural parts of the image, potential objects and object parts; \quad 

\item[($\mathcal{E}_2$)] \emph{People and objects} are detected in the image using YOLO object detection \citep{Redmon2016}; \quad 

\item[($\mathcal{E}_3$)] 
\emph{Human body pose} consisting of body joints and facing direction is extracted using human pose estimation \citep{Cao2017}. 

\end{itemize}

Potential symmetrical structures in the image are defined on the image elements $\mathcal{E}$ using a model of symmetry to identifying pairs of image elements (symmetry pairs) as well as single elements that are constituting a symmetrical configuration.

\smallskip

We consider \emph{compositional structure} {\small\textbf{({\color{black}C1})}} of images, and \emph{similarity} {\small\textbf{({\color{black}C2})}} of constituent elements, in particular perceptual similarity in the low-level features, and semantic similarity of objects and regions. 
The resulting model of symmetrical structure in the image consists of a set of image elements, and the pair-wise similarity relations between the elements.

\subsubsection*{({\bf C1}) Compositional Structure}

Symmetrical composition in the case of reflectional symmetry consists of symmetrically arranged pairs of image elements, where one element is on the left and one is on the right of the symmetry axis, and single centred image elements, which are placed on the symmetry axis. To model this, we represent the extracted image elements as spatial entities, i.e. \emph{points}, \emph{axis-aligned rectangles}, and \emph{line-segments}  and define constraints on the spatial configuration of the image elements, using the following \emph{spatial properties} of the spatial entities:

\begin{itemize}
\item { \emph{position: } the centre-point of a rectangle or position of a point in $x, y$ coordinates;}
\item{ \emph{size: } the width and height of a rectangle $w, h$; }
\item { \emph{aspect ratio: } the ratio $r$ between width and height of a rectangle; }
\item { \emph{distance: }  euclidian distance $d$ between two points $p$ and $q$; }
\item { \emph{rotation: }  the $yaw, pitch,$ and $roll$ angles between two line-segments in 3D space.}
\end{itemize}

\medskip

\emph{Symmetrical Spatial Configuration}\quad
We use a set of spatial relations holding between the image elements to express their spatial configuration; spatial relations (e.g., $left$, $right$, and $on$)\footnote{ The semantics of spatial relations is based on specialised polynomial encoding as suggested in \cite{DBLP:conf/cosit/BhattLS11} within constraint logic programming (CLP) \citep{jaffar1994constraint}; CLP is also the  framework being used to demonstrate Q/A later in this section.} holding between points and lines describe the relative orientation of image elements with respect to the symmetry axis. 
Towards this, we use the relative position ($\operatorname{rel-pos}$) of an image element with respect to the symmetry axis, which is defined as the distance to the symmetry axis and the y coordinate of the element.

\medskip

\emph{--- Image Patches and Objects}\quad  Symmetrical configuration of image elements is defined based on their spatial properties using the following two rules.

\smallskip

In the case of a single element $e$  the centre of the rectangle has to be on the symmetry axis.

\noindent\begin{minipage}{\columnwidth}
{\small
\begin{align}
\begin{split}
&{\predThF{symmetrical}}(e) \supset  \predThF{orientation}(on, \predThF{position}(e), \operatorname{symmetry-axis}). \\
\end{split}
\end{align}}
\end{minipage}

In the case of pairs of elements $e_i$ and $e_j$ these have to be on opposite sites of the symmetry axis, and have same size and aspect ratio, further the position of $e_i$ and $e_j$ has to be reflected.

\noindent\begin{minipage}{\columnwidth}
{\small
\begin{align}
\begin{split}
&{\predThF{symmetrical}}(p_i, p_j) \supset \\
&\quad\quad \predThF{orientation}(left, \predThF{position}(p_i), \operatorname{symmetry-axis})\wedge \\
&\quad\quad \predThF{orientation}(right, \predThF{position}(p_j), \operatorname{symmetry-axis})\wedge \\
&\quad\quad \predThF{equal}(\predThF{aspect-ratio}(p_i),\predThF{aspect-ratio}(p_j))\wedge \\
&\quad\quad \predThF{equal}(\predThF{size}(p_i),\predThF{size}(p_j))\wedge \predThF{equal}(\predThF{rel-pos}(p_i), \predThF{rel-pos}(p_j)).
\end{split}
\raisetag{35pt}
\end{align}}
\end{minipage}

The model of symmetry serves as a basis for analysing symmetrical structures and defines the attributes that constitute a symmetrical configuration. 
Additionally to this basic definition of symmetrical configuration of arbitrary image elements, we define rules for symmetry in the placement and layout of humans  in the images.

\begin{figure}[t]
	\centering
	
	\includegraphics[width=.4\linewidth]{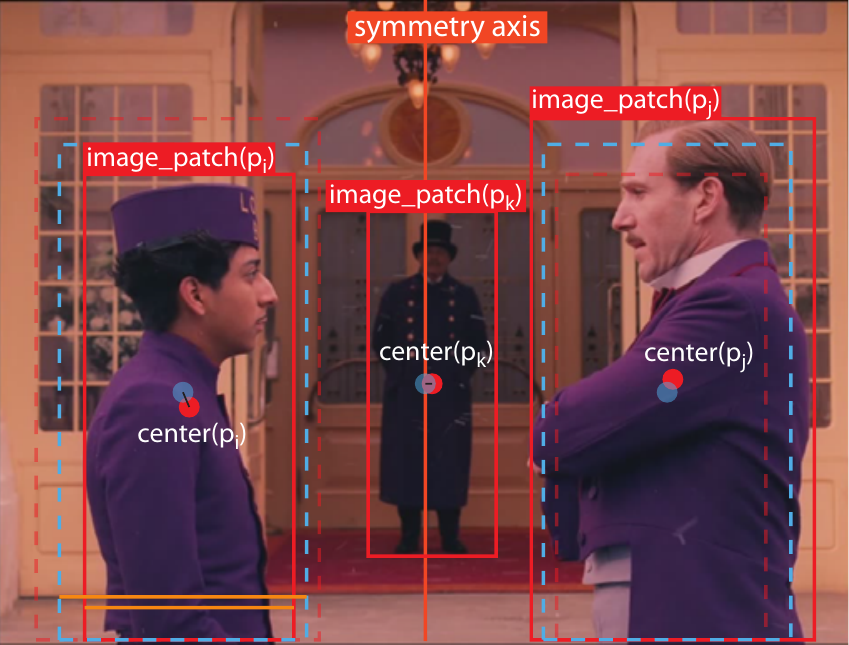}\quad
    	\includegraphics[width=.4\linewidth]{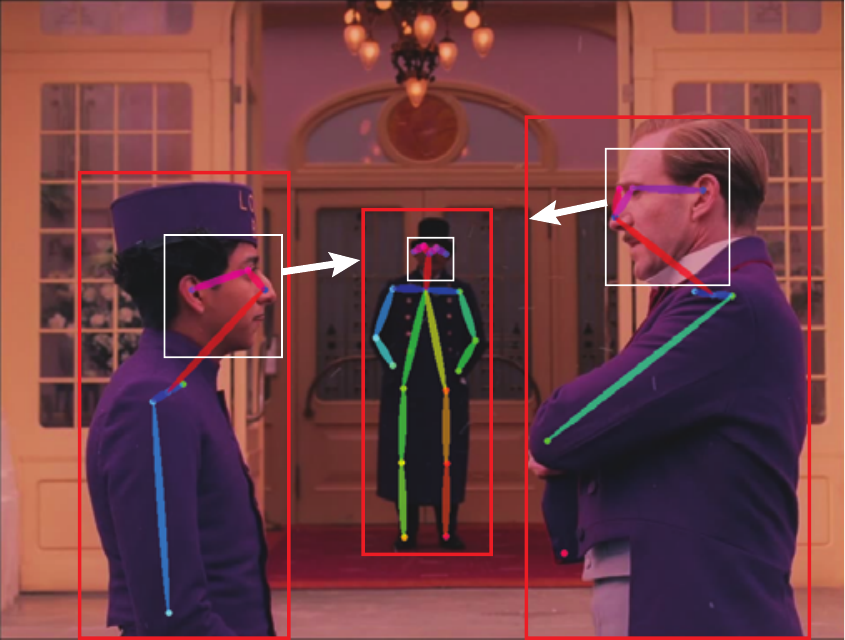}

	\caption{{ Symmetric composition for pairs of image patches, and centering  of single image patches.}} 	
	\label{fig:symmetry_composition}
\end{figure}

\medskip

\emph{--- Human Body Pose}\quad is given by a set of joints $j$, represented as points, i.e. {\small$pose = \{j_0, ..., j_n\}$}. 
The pose can be either symmetrical within itself, or two people can be arranged in a symmetrical way.  
Symmetrical body pose is analysed by defining joint pairs {\small$JP = \{ (j_k, j_l), ..., (j_m, j_n)\}$}, such as {\small$(left\ shoulder, right\ shoulder)$, $(left\ elbow, right\ elbow)$}, etc. and compare the relative position of these pairs with respect to the centre of the person $c_p$.

\noindent\begin{minipage}{\columnwidth}
{\small
\begin{align}
\begin{split}
&{\predThF{symmetrical}}(\predThF{pose}(p)) \supset  \forall (j_k, j_l) \ \predThF{equal}(\predThF{rel-pos}(j_k,  c_p), \predThF{rel-pos}(j_l,  c_p))
\end{split}
\end{align}}
\end{minipage}

Accordingly, pose of two persons is analysed by defining joint pairs associating each joint of one person to the corresponding joint of the other person, e.g., the left hand of person 1 gets associated to the right hand of person 2.

\smallskip

Further we define symmetrical facing directions of two people based on the rotation of their heads. Towards this we use the $yaw, pitch,$ and $roll$ angles of a persons head $h_p$, relatively to a front facing head, and define that the facing direction is  symmetrical if the $pitch$ rotation is the same, and the $yaw$, and $roll$ rotation are opposite.

\noindent\begin{minipage}{\columnwidth}
{\small
\begin{align}
\begin{split}
&{\predThF{symmetrical}}(\predThF{facing\_dir}(p_1), \predThF{facing\_dir}(p_2)) \supset  \\
&\quad\quad \predThF{equal}(\predThF{pitch}(h_{p_1}), \predThF{pitch}(h_{p_2})) \wedge \predThF{equal}(\predThF{yaw}(h_{p_1}), -\predThF{yaw}(h_{p_2}) \wedge  \predThF{equal}(\predThF{roll}(h_{p_1}), -\predThF{roll}(h_{p_2})).
\end{split}
\end{align}}
\end{minipage}

\medskip

\emph{Divergence from Symmetrical Configuration}\quad
%
To account for configurations that are only symmetrical in some aspects, as it typically occurs in naturalistic scenes, we calculate the divergences of the configuration from the symmetry model.
For each element of the symmetry structure we calculate the divergence from the defined symmetry model, i.e., we focus on divergence with respect to position, size, aspect ration, and pose (involving configuration of body parts and joints). We use thresholds on the average of these values to identify hypotheses on (a)symmetrical structures.


\subsubsection*{({\bf C2}) Similarity Measures}

Visual Symmetry is also based on similarity of image features; we assess similarity of image patches using CNN features, e.g, obtained from AlexNets \citep{krizhevsky2012imagenet}, or ResNets \citep{He2016_resnet}, pre-trained on the ImageNet Dataset \citep{imagenet2009}, i.e., we use the extracted features to evaluate perceptual similarity and use ImageNet classifications to evaluate semantic similarity of image patches.


\medskip

\emph{Perceptual Similarity} \quad
%
Visual Symmetry is based in perceptual similarity of image features, this denotes the similarity in visual appearance of the image patches. 
To analyse the perceptual similarity of image patches we use the feature vectors obtained from the network and use cosine similarity to evaluate the similarity of the feature vectors of two image patches.
For the case of reflectional symmetry we compare the image patches of all potential symmetry pairs by comparing the features of one patch to the features of the mirrored second patch.


\medskip

\emph{Semantic Similarity} \quad
%
On the semantical level, we classify the image patches and compare their content for semantic similarities, i.e. we compare conceptual similarity of the predicted categories.
Towards this we use the weighted ImageNet classifications for each image patch with WordNet \citep{Miller1995wordnet}, which is used as an underlying structure in ImageNet, to estimate conceptual similarity of the object classes predicted for the image patches in each symmetry pair. In particular, we use the top five predictions from the AlexNet classifiers and estimate similarity of each pair by calculating the weighted sum of the similarity values for each pair of predicted object categories.

\begin{table}
\begin{center}
\scriptsize
\begin{tabular}{|l|p{2.7in}|}
\hline
{\textbf{Predicate}} & \textbf{Description} \\\hline

 &\\

{{\sffamily symmetrical\_element}($E$)} & Symmetrical elements $E$.\\[2pt]

{{\sffamily non\_symmmetrical\_element}($E$)} & Non-symmetrical elements $E$.\\[2pt]

{{\sffamily symmetrical\_objects}($SO$)} & Symmetrical objects $SO$.\\[2pt]

{{\sffamily non\_symmetrical\_objects}($NSO$)} & Non-symmetrical objects $NSO$.\\[2pt]

{{\sffamily symmetrical\_body\_pose}($SP$,$SBP$)} & Symmetrical person $SP$ (pair or single object), and symmetrical parts of body-pose $SBP$.\\[2pt]

{{\sffamily non\_symmetrical\_body\_pose}($SE$,$NSP$)} & Symmetrical person $SP$ (pair or single object), and non-symmetrical parts of body-pose $SBP$.\\[2pt]

{{\sffamily symmetry\_stats}($NP, NSP, MD, MS$)} & Basic stats on symmetrical structure: number of patches $NP$, number of symmetrical patches $NSP$, mean divergence $MD$, and mean similarity $MS$.\\[2pt]

{{\sffamily symmetrical\_objects\_stats}($NO, NSO, MD, MS$)} & Stats on symmetrical structure of objects: number of objects $NO$, number of symmetrical objects $NSO$, mean divergence $MD$, and mean similarity $MS$.\\

&\\\hline


\end{tabular}
\normalsize
\caption{Sample predicates for querying interpretation model.}
\label{tbl:sym_preds}
\end{center}
\end{table}%

\subsection{Declarative Symmetry Semantics }\label{sec:declarative_model}
The semantic structure of symmetry is described by the model in terms of a set of symmetry pairs and their respective similarity values with respect to the three layers of our model, i.e. for each symmetry pair it provides the similarity measures based on semantic similarity, spatial-arrangement, and low-level perceptual similarity (Table \ref{tbl:computational_steps}). This results in a declarative model of symmetrical structure, which is used for fine-grained analysis of symmetry features and question-answering about symmetrical configuration in images, i.e., using our framework, it is possible to define high-level rules and execute queries in (constraint) logic programming \citep{jaffar1994constraint} (e.g., using SWI-Prolog \citep{wielemaker:2011:tplp}) to reason about symmetry and directly \emph{query}  symmetrical features of the image.\footnote{Within the (constraint) logic programming language \textsc{Prolog}, ` \textbf{,} ' corresponds to conjunction, ` \textbf{;} ' to a disjunction, and `a \textbf{:-} b, c.' denotes a \emph{rule} where `a' is true if both `b' and `c' are true; capitals are used to denote variables, whereas lower-case refers to constants; `\textbf{\underline{\hspace{5pt}}}' (i.e., the underscore) is a ``dont care'' variable, i.e., denoting placeholders for variable in cases where one doesn't care for a resulting value.}

\begin{table}
\centering
\scriptsize
\begin{tabular}{|p{0.95\textwidth}|}

\hline

\vspace{-2pt}

\textbf{Step 1) } \textit{Extracting Image Elements} \quad
extract image elements $\mathcal{E}$ consisting of $\mathcal{E}_1,\mathcal{E}_2, and $~$\mathcal{E}_3$ : 

\vspace{2pt}
($\mathcal{E}_1$) \emph{Image Patches} are extracted using selective search as described in \cite{Uijlings2013};  

\vspace{2pt}
($\mathcal{E}_2$) \emph{People and Objects} are detected in the image using YOLO object detection \citep{Redmon2016};  

\vspace{2pt}
($\mathcal{E}_3$) 
\emph{Human Body Pose} consisting of body joints and facing direction is extracted using human pose estimation \citep{Cao2017}.

\medskip

\centerline{
	\includegraphics[width=0.3\columnwidth]{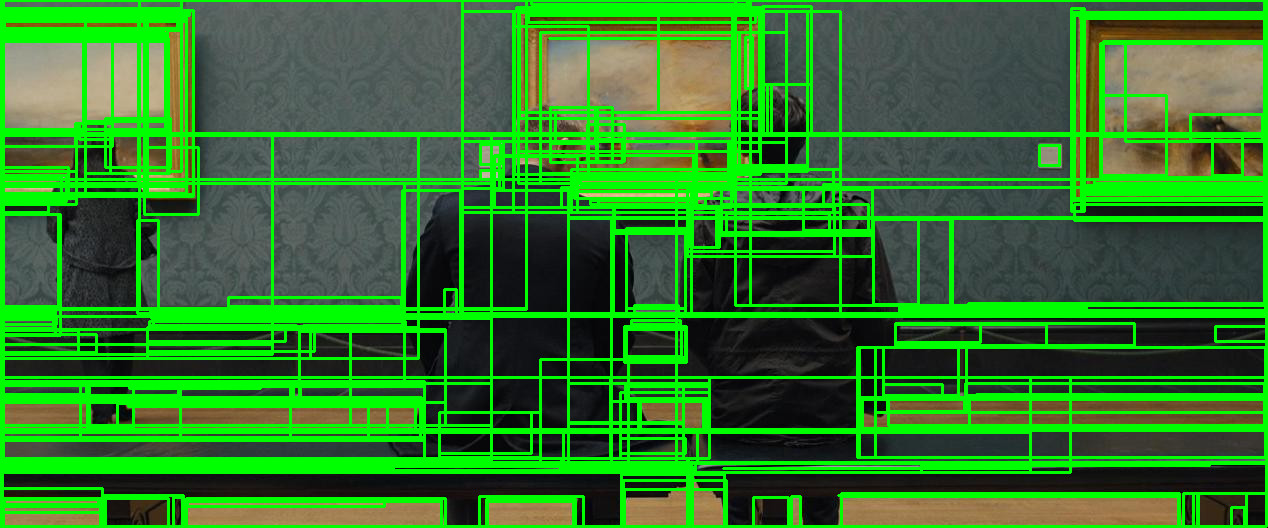}\hfill
	\includegraphics[width=0.3\columnwidth]{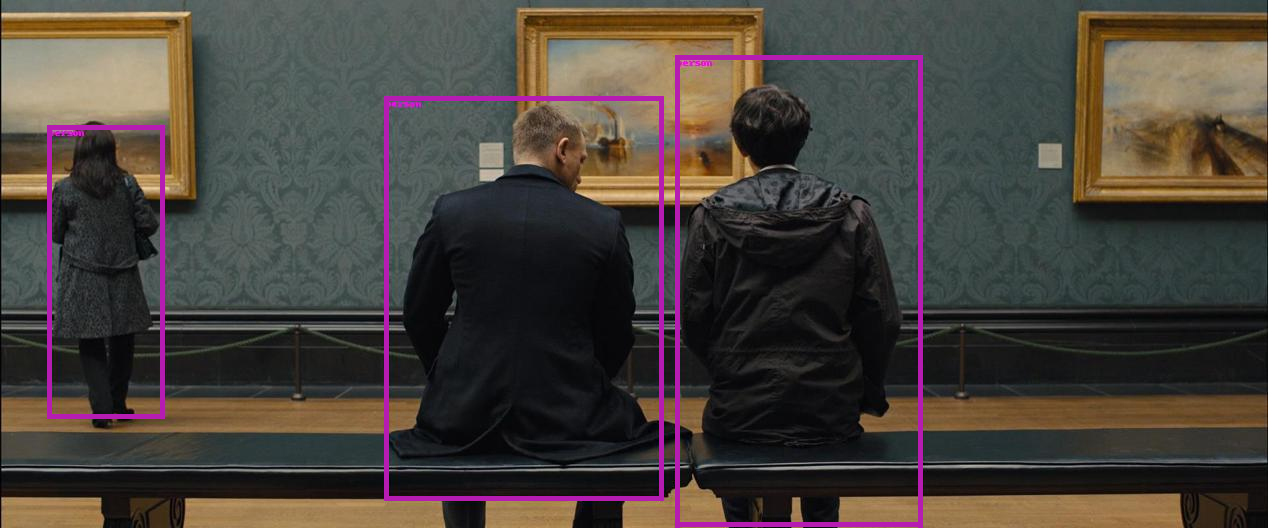}\hfill
	\includegraphics[width=0.3\columnwidth]{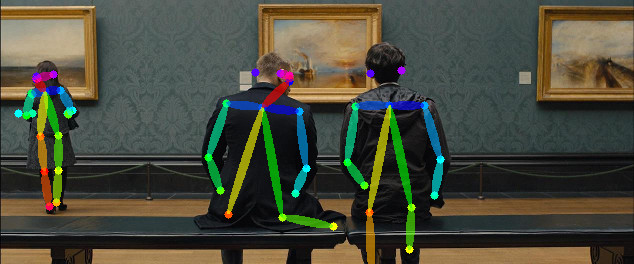}
}

\medskip

\includegraphics[width=\linewidth]{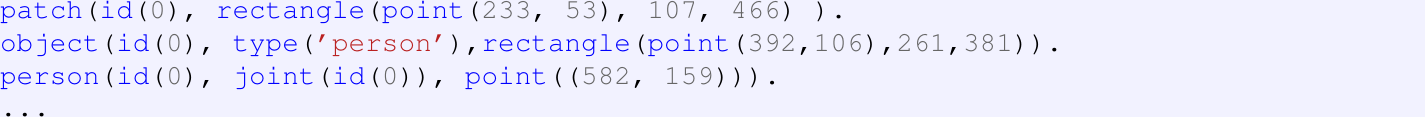}
%
%
\\

\begin{minipage}[c]{0.65\linewidth}


\vspace{4pt}

\textbf{Step 2) } \textit{Semantic and Perceptual Similarity} \quad 
Compute semantic and perceptual similarity for each pair of image elements  $e_i$ and $e_j \in \mathcal{E}$ based on features from CNN layers. \\[2pt]
-- compute \emph{semantic similarity} based on ImageNet classification of image patches

-- compute \emph{perceptual similarity} based on cosign similarity between CNN features\\ $~$ $~$  of image elements
\end{minipage}
\hfill
\begin{minipage}[c]{0.3\linewidth}
\centerline{\includegraphics[width=0.9\columnwidth]{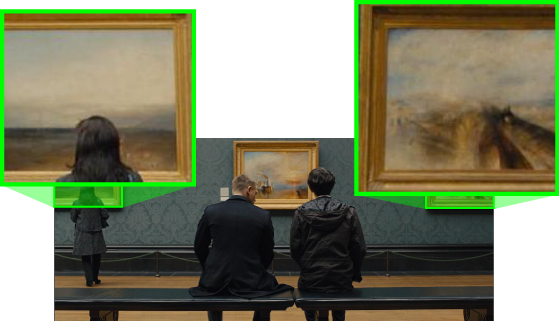}
}
\end{minipage}

\smallskip

\includegraphics[width=\linewidth]{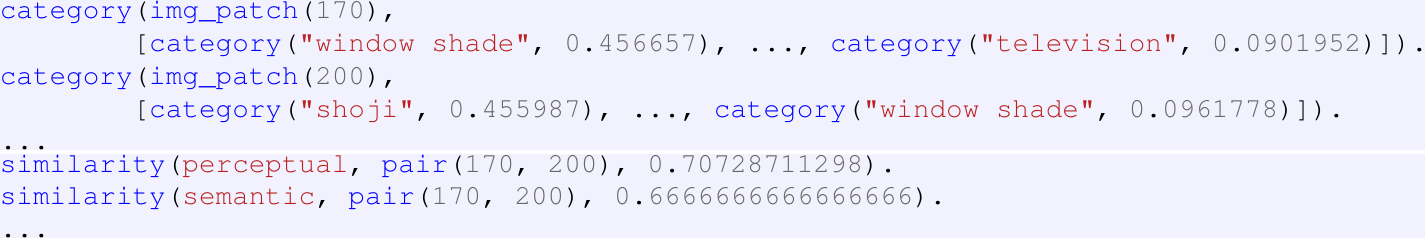}
%
%
\\[4pt]


\textbf{Step 3) }  \textit{Symmetry Configuration and Divergence} \quad 
Identify symmetrical structures in the image elements $\mathcal{E}$ based on the formal definition of symmetry in Section \ref{sec:reflectionsym} and calculate the divergence of elements from this model. 

\smallskip

\includegraphics[width=\linewidth]{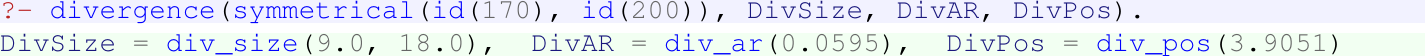}


\medskip
\medskip

\centerline{
\begin{minipage}[c]{0.3\linewidth}
\centerline{\scriptsize\sffamily{Threshold on divergence}}
\end{minipage}\quad\quad
\begin{minipage}[c]{0.05\linewidth}
$~$
\end{minipage}\quad\quad
\begin{minipage}[c]{0.3\linewidth}
\centerline{\sffamily{Threshold on similarity}}
\end{minipage}
}

\centerline{
\begin{minipage}[c]{0.3\linewidth}
\includegraphics[width=\columnwidth]{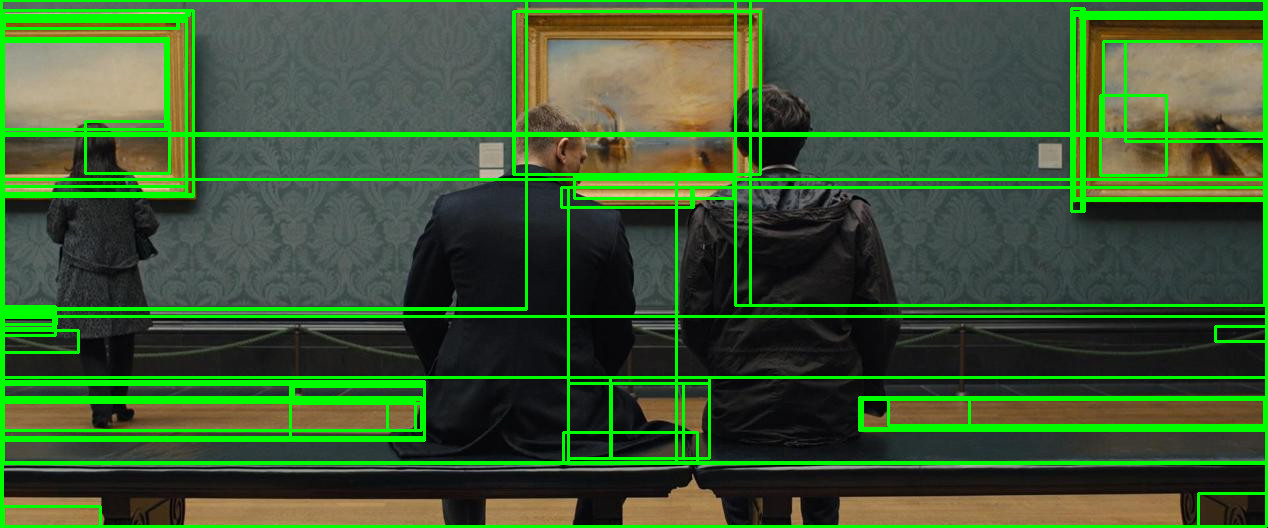}
\end{minipage}\quad\quad
\begin{minipage}[c]{0.05\linewidth}
\includegraphics[width=\columnwidth]{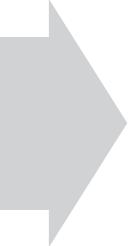}
\end{minipage}\quad\quad
\begin{minipage}[c]{0.3\linewidth}
\includegraphics[width=\columnwidth]{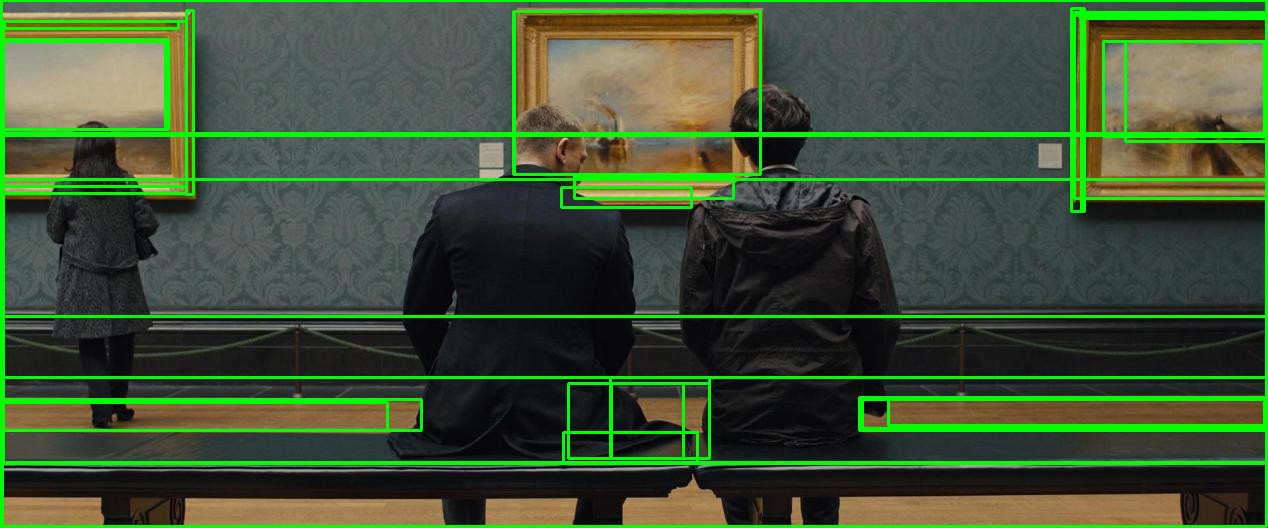}
\end{minipage}
}
\\\hline\hline

\vspace{-2pt}

\textbf{Result } \textit{A Declarative Model for Semantic Analysis of Symmetry} \quad
The process results in the declarative structure consisting of the symmetrical properties of the image given by the elements of the image $\mathcal{E}$ and the divergence from the formal definition of symmetry. This model serves as a basis for declarative query-answering about symmetrical characteristics of the image, e.g. the images showcase results from queries in Section \ref{sec:declarative_model}, analysing symmetrical an non-symmetrical image elements.

\medskip

\centerline{
\begin{minipage}[c]{0.45\linewidth}
\centerline{\scriptsize\sffamily{Symmetrical Elements}}
\end{minipage}$~$$~$
\begin{minipage}[c]{0.45\linewidth}
\centerline{\sffamily{Non-Symmetrical Elements}}
\end{minipage}
}

\centerline{
\begin{minipage}[c]{0.45\linewidth}
	\includegraphics[width=\columnwidth]{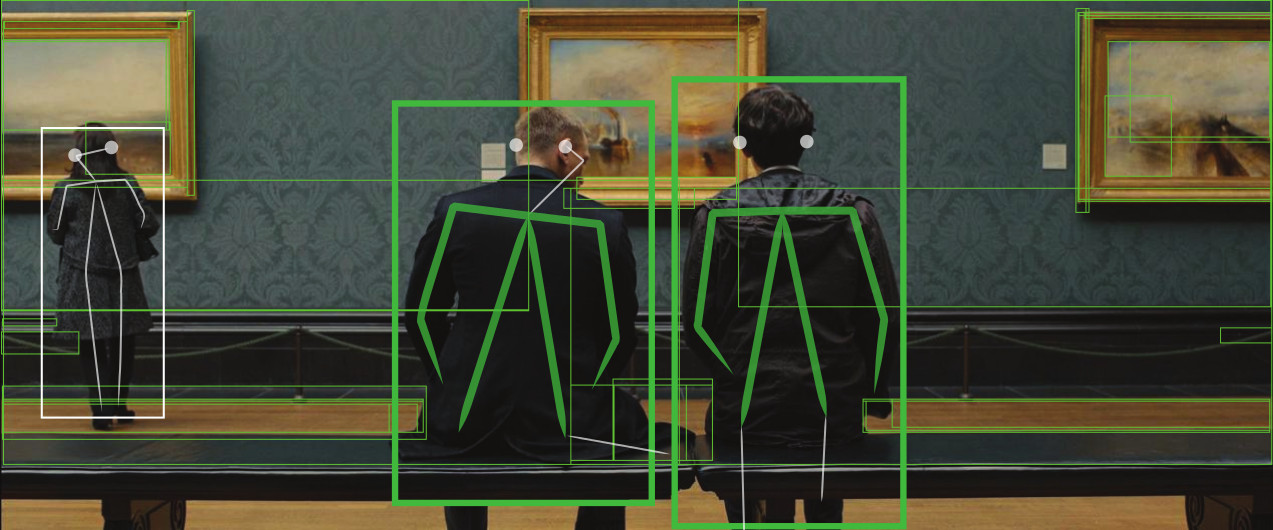}
\end{minipage}$~$$~$
\begin{minipage}[c]{0.45\linewidth}
	\includegraphics[width=\columnwidth]{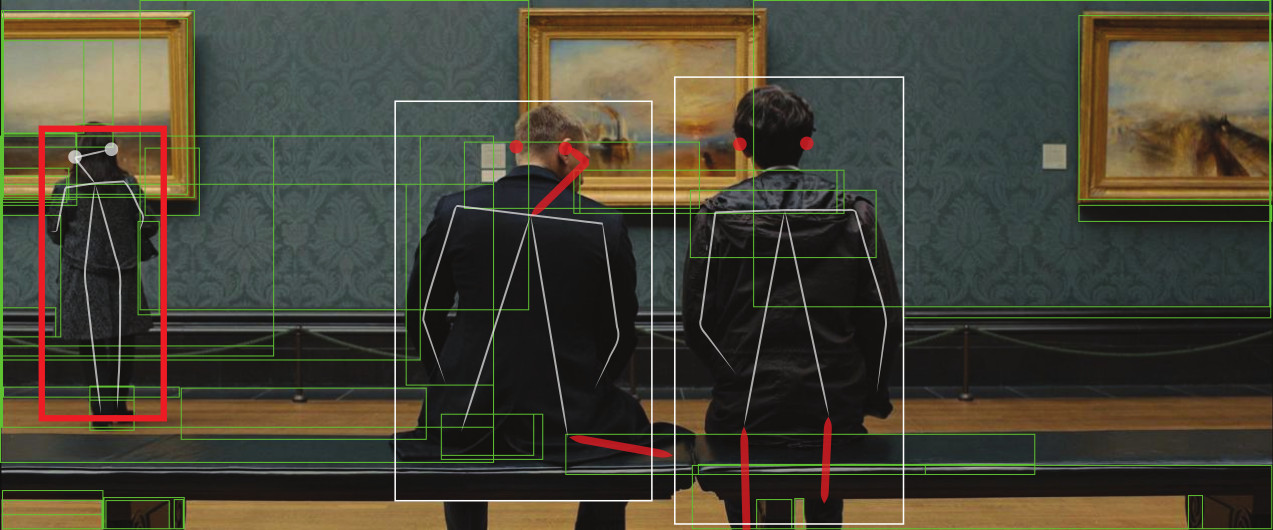}
\end{minipage}
}

\\\hline

\end{tabular}
\caption{Computational steps to generate the semantic symmetry model. }
\label{tbl:computational_steps}
\end{table}

\paragraph{Symmetrical Structure of the Image}

As an example consider the image in Table \ref{tbl:computational_steps}.  
Based on the symmetrical structure extracted from the image, the underlying interpretation model is queryable using utility predicates (see sample predicates in Table \ref{tbl:sym_preds}). 
The symmetry model as defined in Section \ref{sec:reflectionsym} 
can be used to query symmetrical (and non-symmetrical) elements of the image using the following rules:

\smallskip

\includegraphics[width=\linewidth]{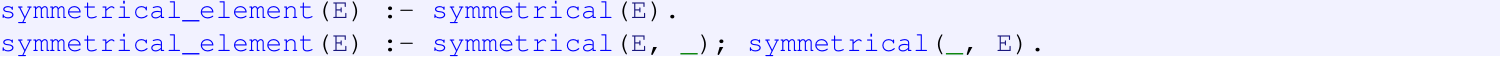}

%
%

\smallskip

{Aggregating results for the {\small{\sffamily symmetrical\_element}($E$)} predicate for the example image results in a list of all symmetrical image elements (depicted in the results in Table \ref{tbl:computational_steps}):}

\smallskip

\includegraphics[width=\linewidth]{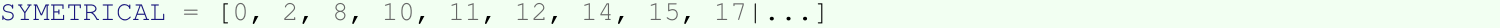}

%
%
%

\smallskip

Similarly we can query for the non symmetrical elements of the image using the following rule:

\smallskip

\includegraphics[width=\linewidth]{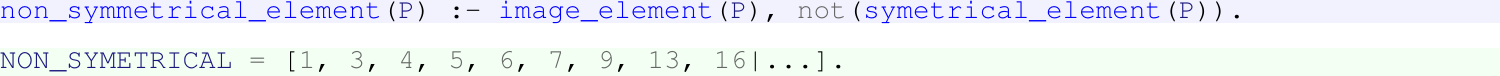}

%
%
%
%
%
%

\medskip

\textit{Divergence} \quad 
The divergence of a image elements from the optimal symmetrical configuration can be directly queried using the {\sffamily\footnotesize divergence} predicate:

\smallskip

\includegraphics[width=\linewidth]{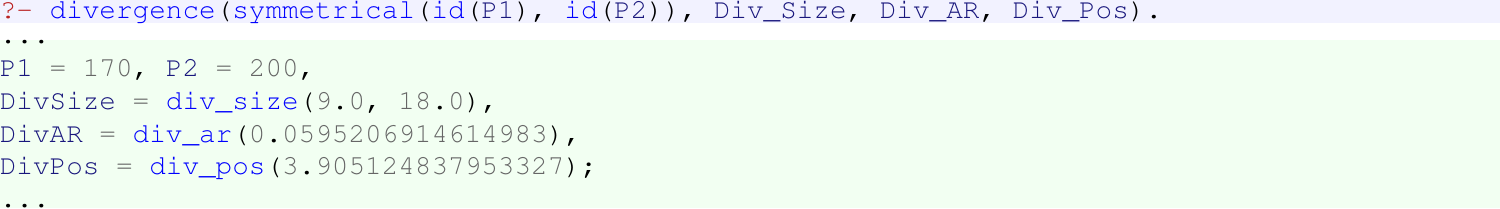}

%
%

%
%

\medskip

\textit{Similarity} \quad Perceptual and semantic similarity of image elements are queried as follows:

\smallskip

\includegraphics[width=\linewidth]{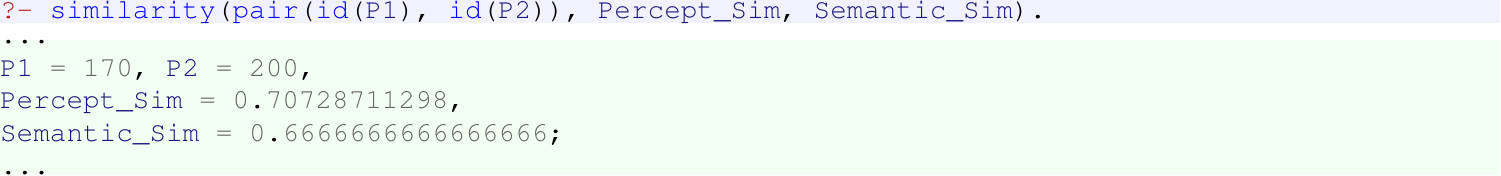}

%
%
%
%

%
%

\smallskip

The above predicates provide the basis for the semantic analysis of symmetry structures in the image as described in the following.

\paragraph{Symmetrical Structure of Objects and People}

Symmetrical structures in the configuration of \emph{objects} and \emph{people} in the image can be queried using the predicat {\sffamily\footnotesize symmetrical\_objects} to get symmetrically configured objects, i.e., pairs of symmetrically positioned objects and single objects in the centre of the image.

\smallskip

\includegraphics[width=\linewidth]{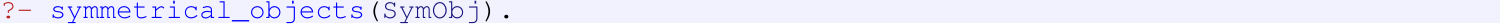}

%
%

\smallskip

For the example image this results in the two people sitting on the bench in the centre of the image.

\smallskip

\includegraphics[width=\linewidth]{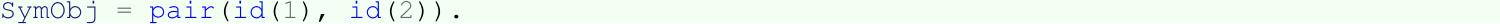}


\smallskip

Similarly to symmetrical object configurations, objects placed in a \emph{non-symmetrical} way can be queried as follows:
 
 \smallskip
 
\includegraphics[width=\linewidth]{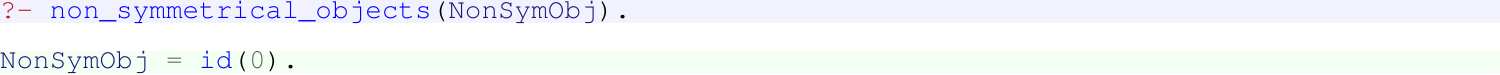} 

%
%
%
%
%
%

\smallskip

Resulting in objects that are not part of a symmetrical structure,
i.e., the person in the left of the image has no symmetrical correspondent in the right of the image.

\medskip

\emph{Body Pose}.\quad 
Based on this, the extracted symmetrical objects can be analysed further, e.g., symmetrical configuration of \emph{people} and their \emph{body pose}, can be queried using the following rule:

\smallskip

\includegraphics[width=\linewidth]{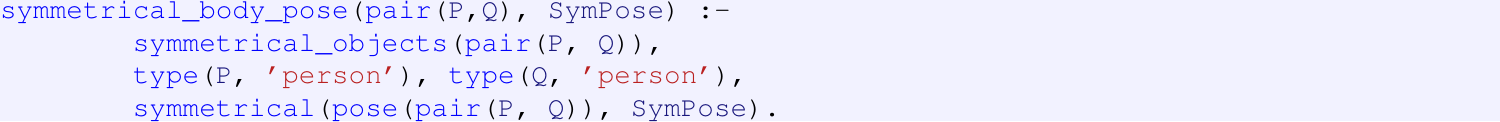}

%

\smallskip

This results in the symmetrically placed people, and the elements of the poses that are symmetrical, i.e., the upper-body of person 1 and person 2 are symmetrical.

\smallskip

\includegraphics[width=\linewidth]{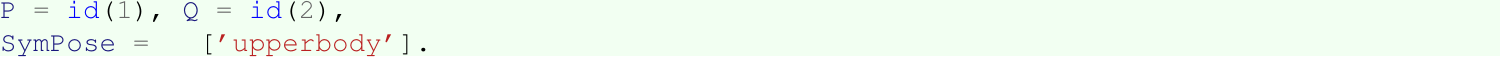}

%
%

\smallskip

Respectively, \emph{non-symmetrical parts} of the \emph{body pose} can be queried as follows:

\includegraphics[width=\linewidth]{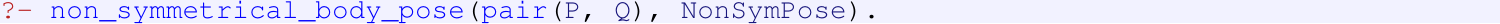}


\smallskip

Resulting in the parts of the body poses that are not symmetrical.

\smallskip

\includegraphics[width=\linewidth]{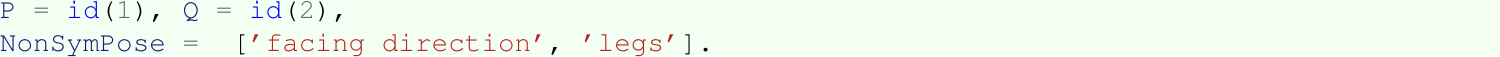}

%
%

\smallskip

As such, the above analysis states that the two people sitting on the bench are placed in a symmetrical way. Their pose is symmetrical in the upper-body, while the facing direction and the legs are not symmetrical.


\begin{table}[t]
\centering
\scriptsize
\begin{tabular}{@{}p{.33\textwidth}@{}p{.22\textwidth}@{}p{.22\textwidth}@{}p{.22\textwidth}@{}}


\vspace{0cm}\multirow{ 3}{*}{\includegraphics[width = .31\textwidth]{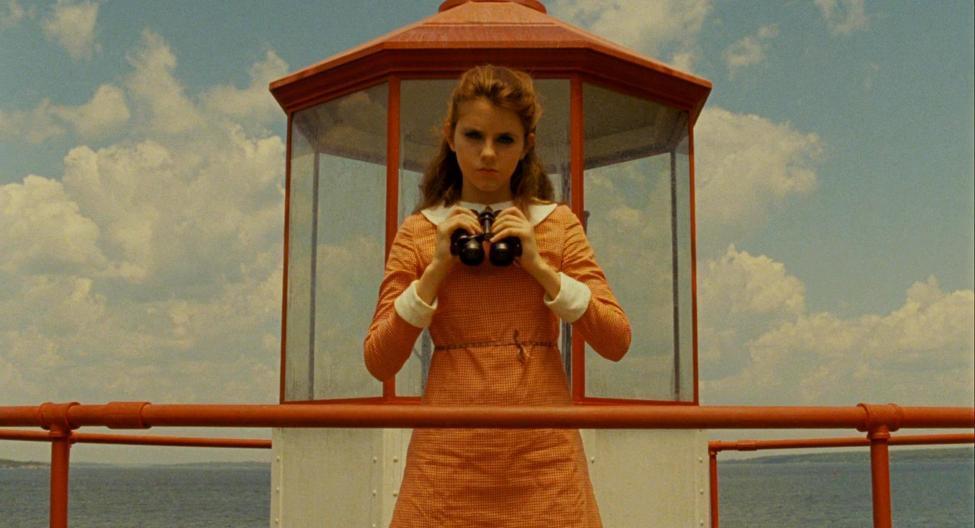}}&
\vspace{0cm}\includegraphics[width = .21\textwidth]{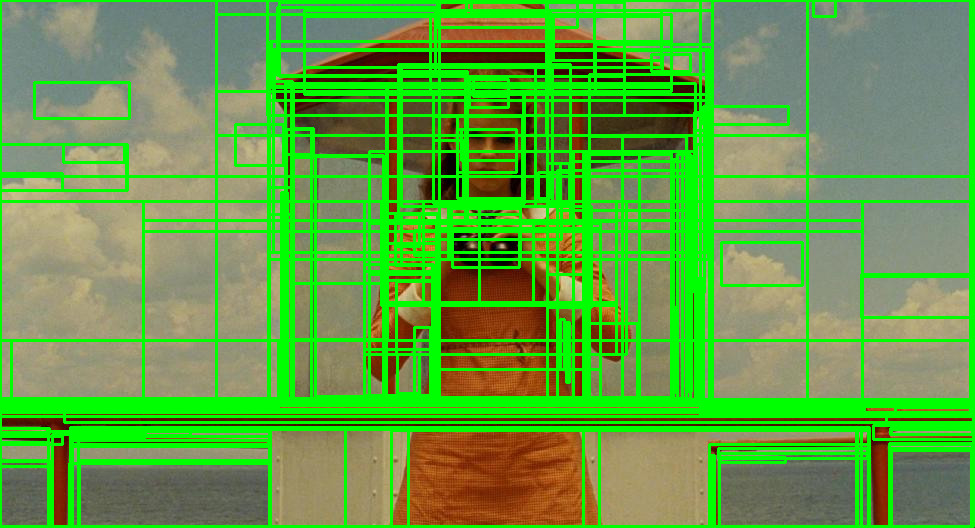}&
\vspace{0cm}\includegraphics[width = .21\textwidth]{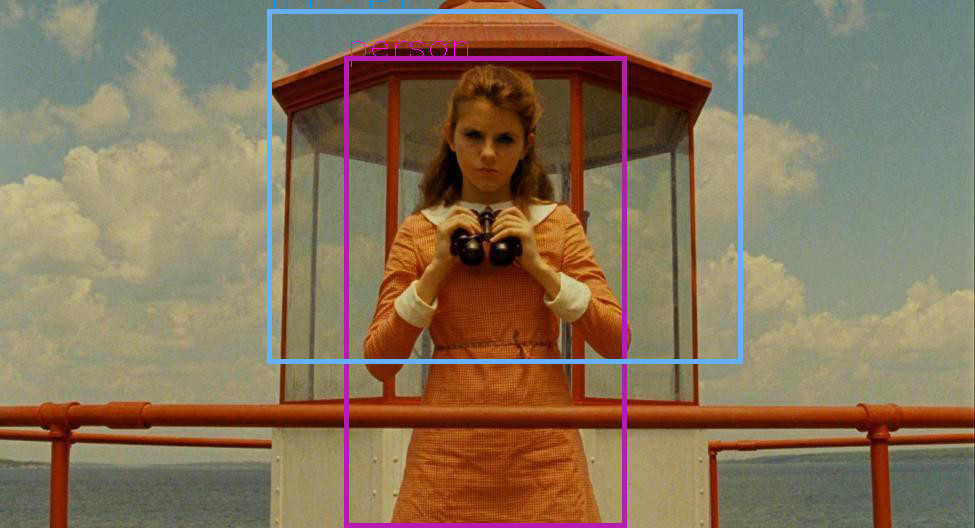}&
\vspace{0cm}\includegraphics[width = .21\textwidth]{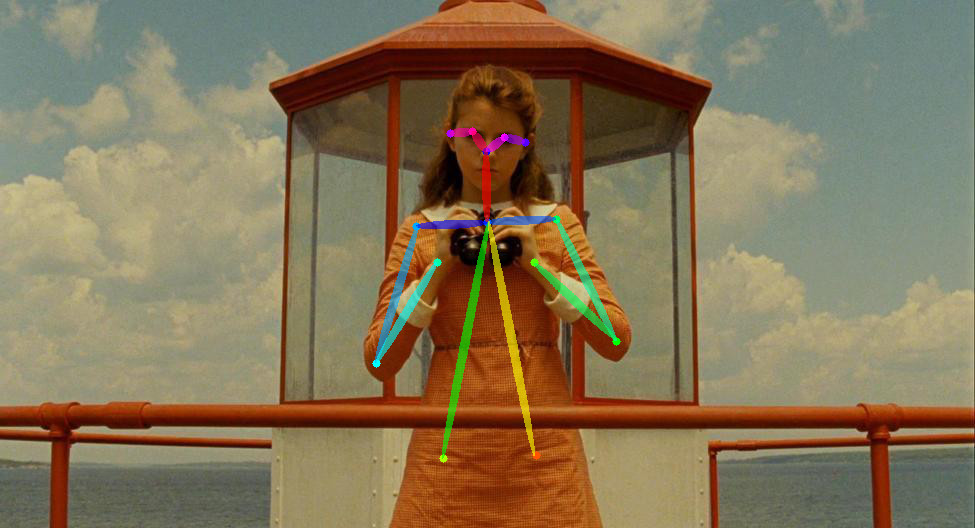}\\[5pt]

 & \multicolumn{3}{c}{\tiny\sffamily Num. Elements: {\bf\sffamily 232} \quad\quad Sym. Elements: {\bf\sffamily 26} \quad\quad Rel. Sym.: {\bf\sffamily 0.112} \quad\quad Mean Perceptual Sim.: {\bf\sffamily 0.813}}\\[7pt]

\vspace{0cm}\multirow{ 2}{*}{\includegraphics[width = .31\textwidth]{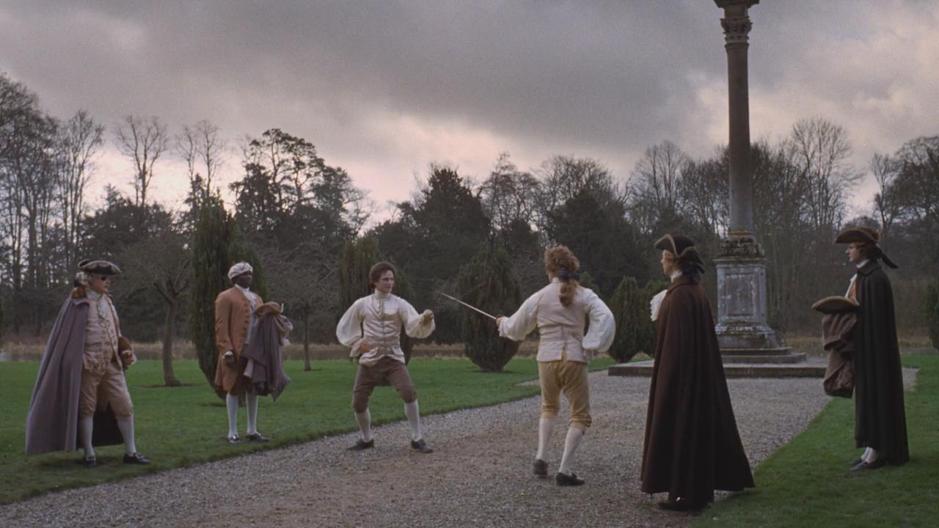}}&
\vspace{0cm}\includegraphics[width = .21\textwidth]{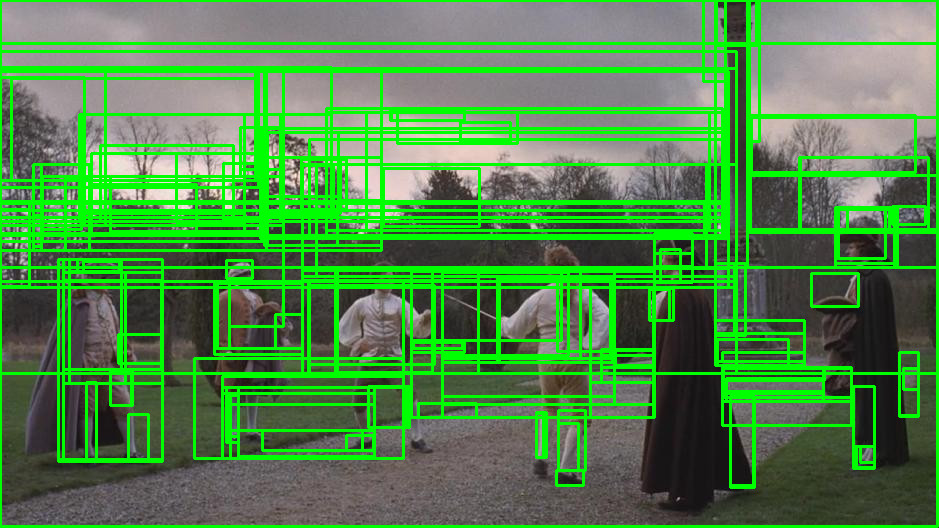}&
\vspace{0cm}\includegraphics[width = .21\textwidth]{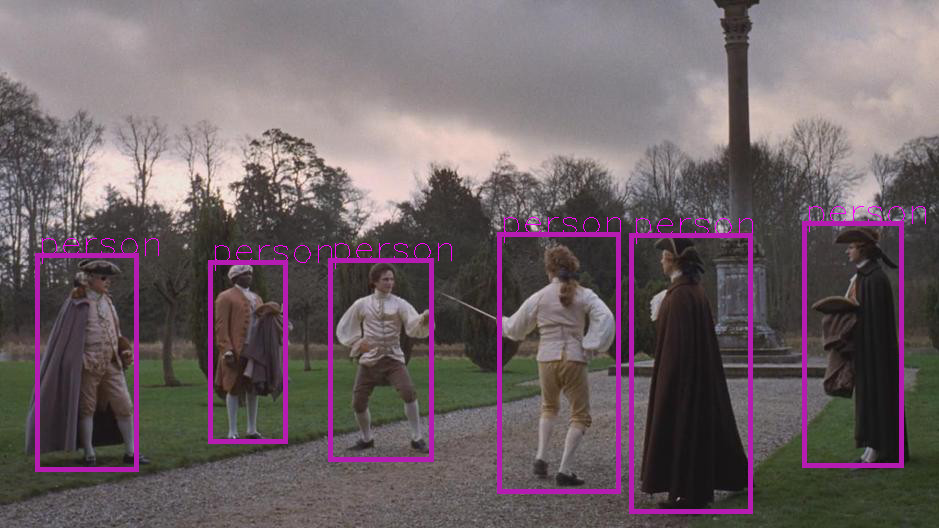}&
\vspace{0cm}\includegraphics[width = .21\textwidth]{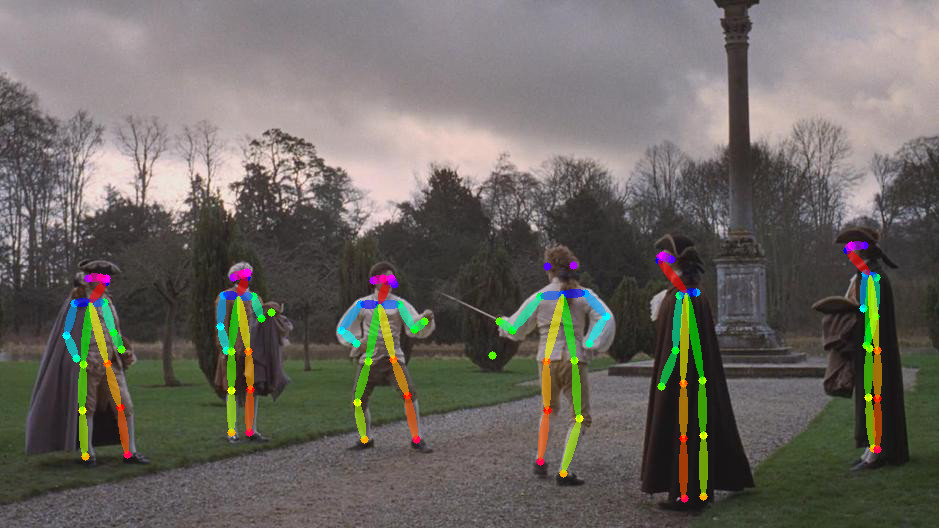}\\[5pt]

 & \multicolumn{3}{c}{\tiny\sffamily Num. Elements: {\bf\sffamily $~$77} \quad\quad Sym. Elements: {\bf\sffamily $~$2} \quad\quad Rel. Sym.: {\bf\sffamily 0.026} \quad\quad Mean Perceptual Sim.: {\bf\sffamily 0.941}}\\[8pt]

\vspace{0cm}\multirow{ 2}{*}{\includegraphics[width = .31\textwidth]{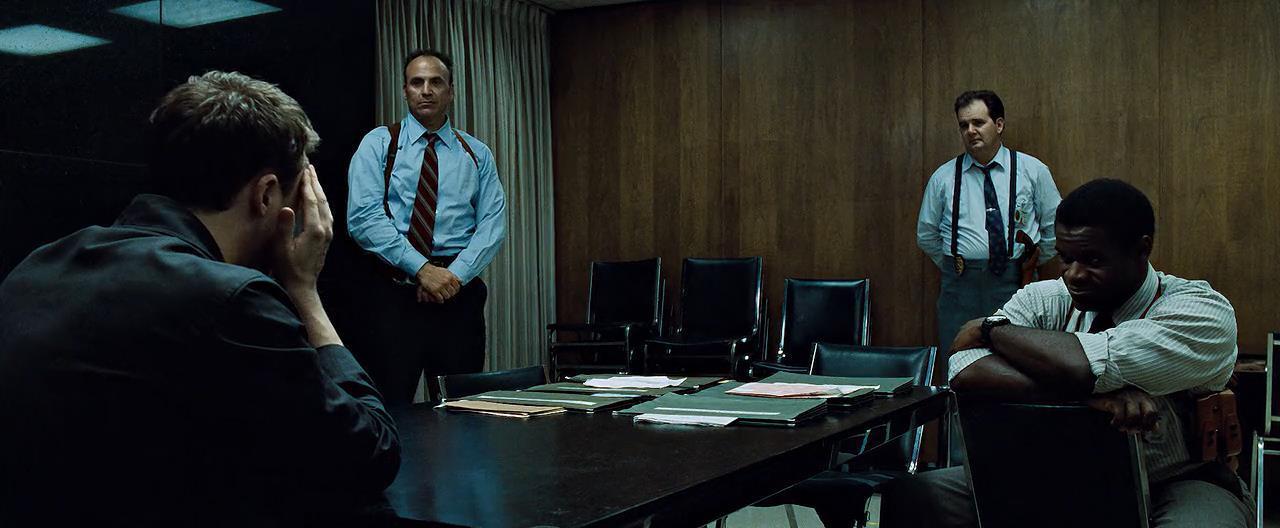}}&
\vspace{0cm}\includegraphics[width = .21\textwidth]{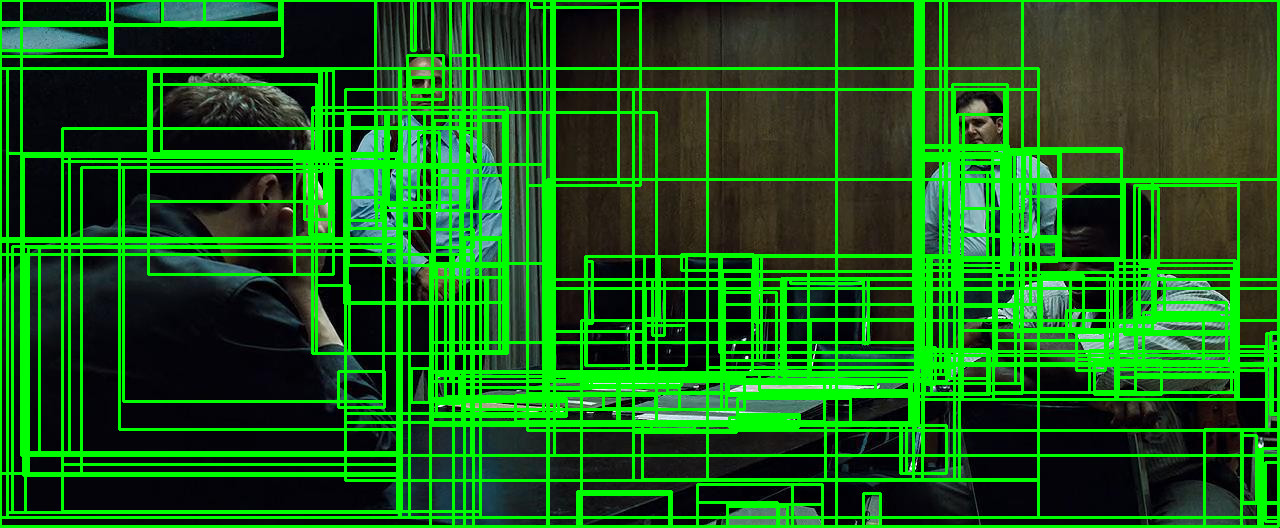}&
\vspace{0cm}\includegraphics[width = .21\textwidth]{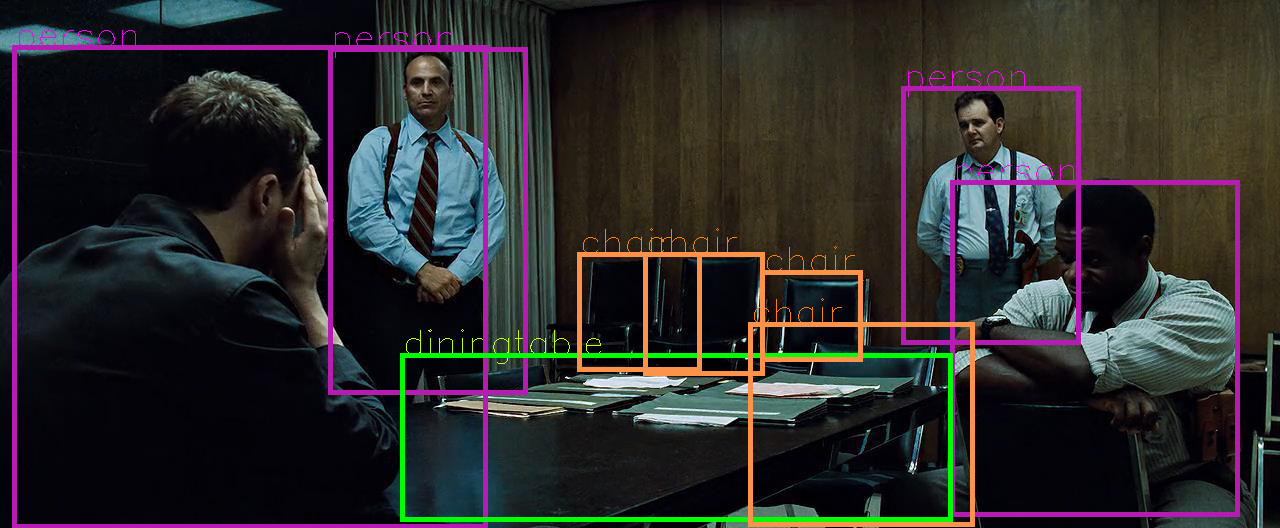}&
\vspace{0cm}\includegraphics[width = .21\textwidth]{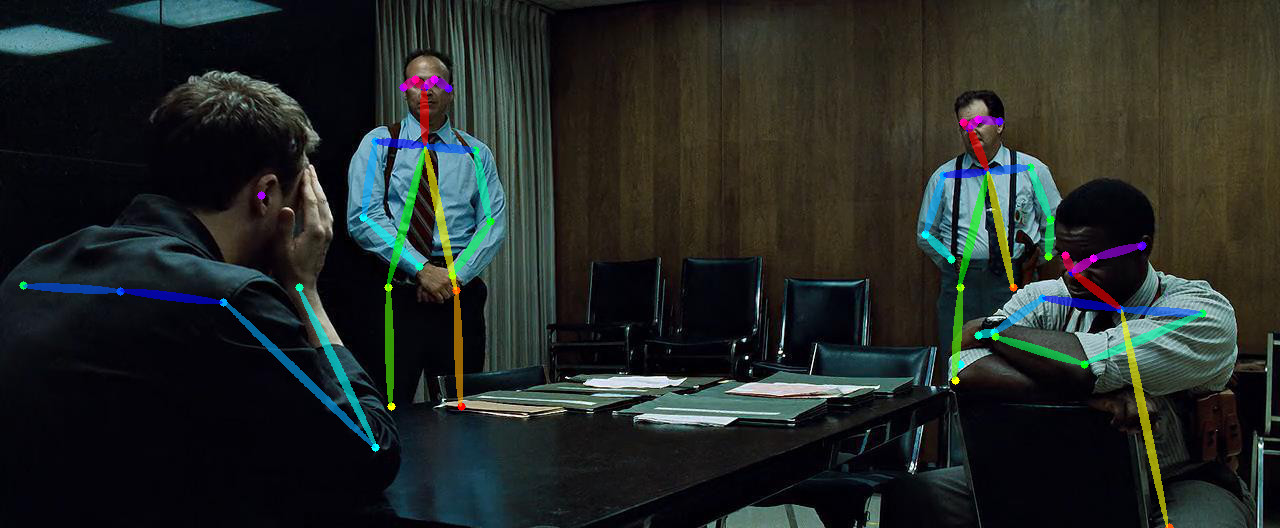}\\[3pt]

 & \multicolumn{3}{c}{\tiny\sffamily Num. Elements: {\bf\sffamily 335} \quad\quad Sym. Elements: {\bf\sffamily $~$5} \quad\quad Rel. Sym.: {\bf\sffamily 0.015} \quad\quad Mean Perceptual Sim.: {\bf\sffamily 0.75$~$}}\\[4pt]

\vspace{0cm}\multirow{ 2}{*}{\includegraphics[width = .31\textwidth]{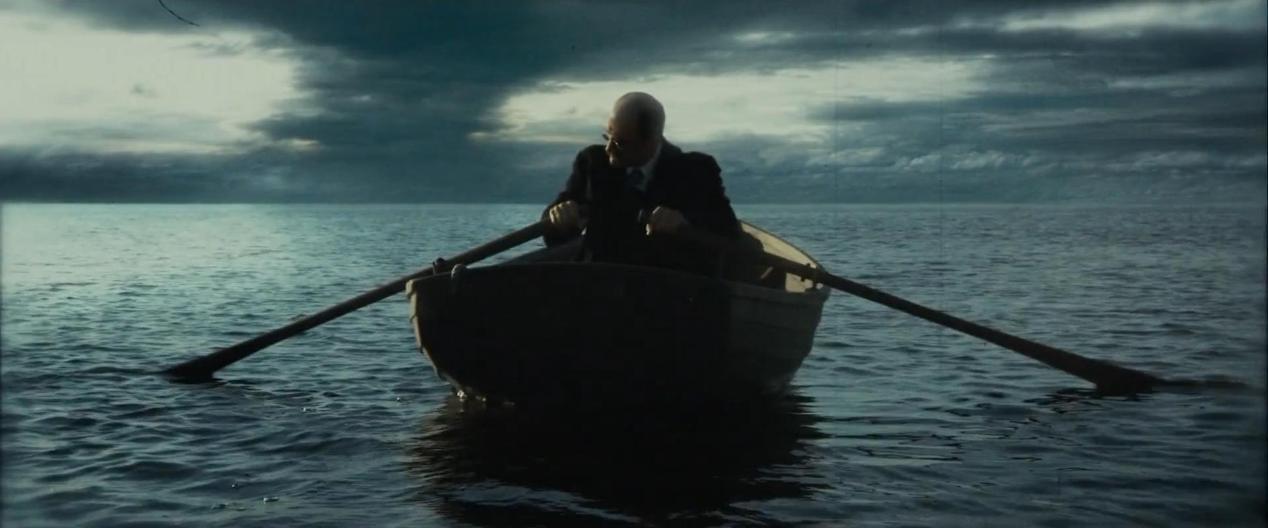}}&
\vspace{0cm}\includegraphics[width = .21\textwidth]{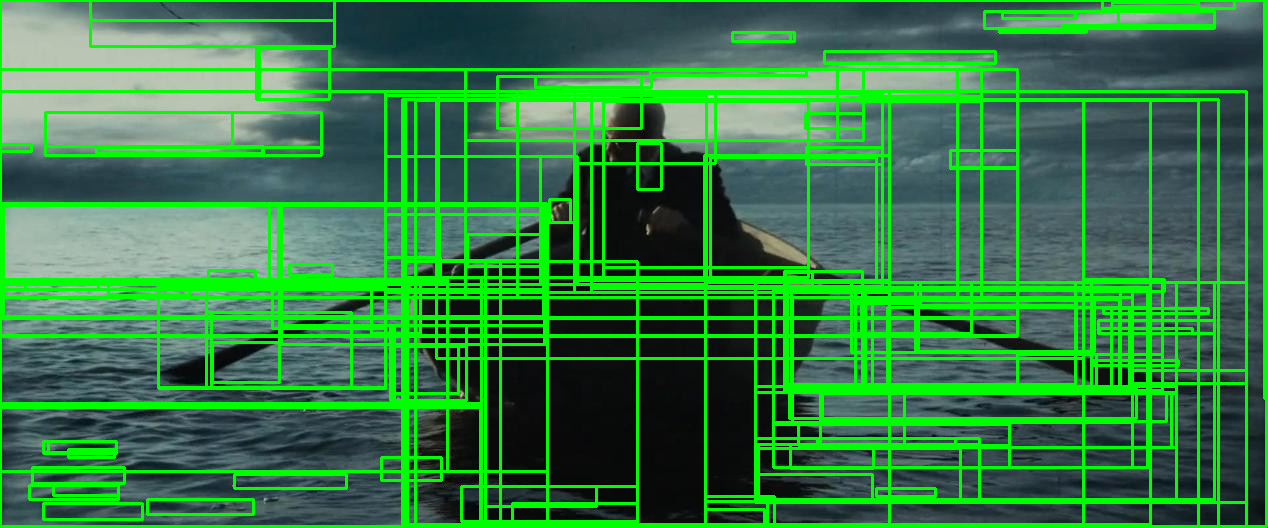}&
\vspace{0cm}\includegraphics[width = .21\textwidth]{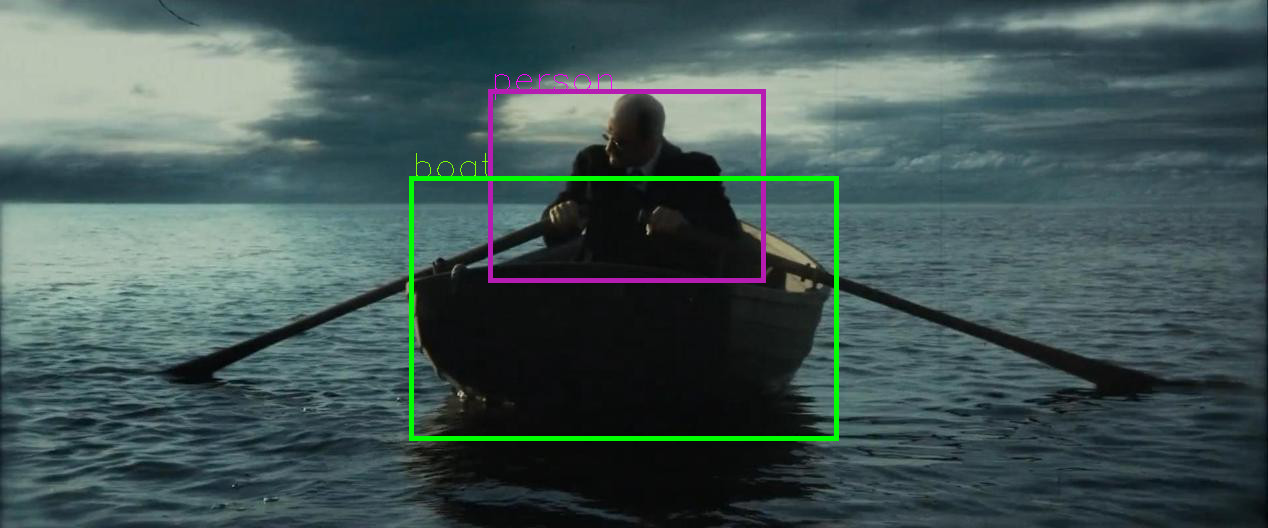}&
\vspace{0cm}\includegraphics[width = .21\textwidth]{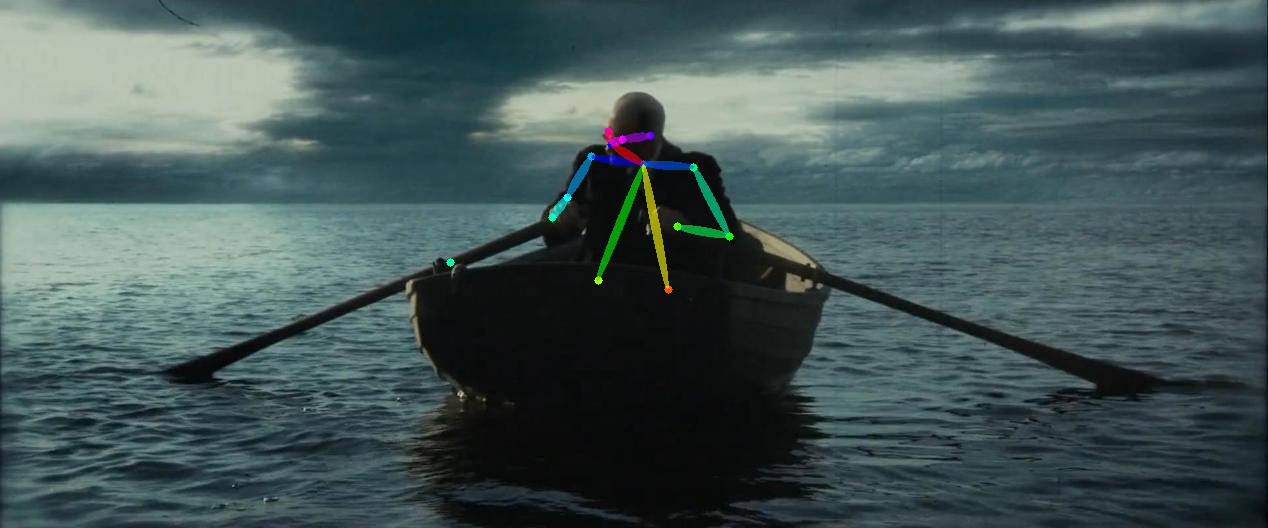}\\[5pt]

 & \multicolumn{3}{c}{\tiny\sffamily Num. Elements: {\bf\sffamily 199} \quad\quad Sym. Elements: {\bf\sffamily $~$8} \quad\quad Rel. Sym.: {\bf\sffamily 0.04$~$} \quad\quad Mean Perceptual Sim.: {\bf\sffamily 0.916}}\\[6pt]

\end{tabular}
\caption{Extracted elements and statistics on symmetry structures for exemplary images.}
\label{tbl:images}
\end{table}

\paragraph{Statistics on Image Symmetry}

Additionally the model can be used to query statistics on the symmetrical features of an image, e.g., to train a a classifier based on the semantic characterisations of symmetry as shown in Section \ref{sec:human-eval}. (see Table \ref{tbl:images} for additional examples of symmetry statistics)

\smallskip

\includegraphics[width=\linewidth]{paper-assets/minted/minted_15.pdf}

%
%
%
%
%
%
%

\smallskip

Similarly statistics on symmetry of objects and people can be queried.

\smallskip

\includegraphics[width=\linewidth]{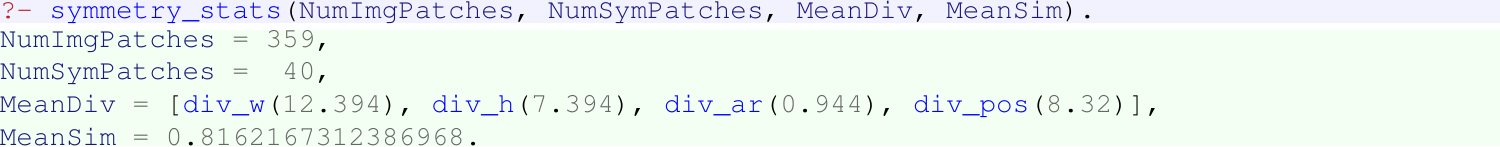}

%
%
%
%
%
%

\medskip
\medskip

Based on these rules, our model provides a declaratively interpretable characterisation of reflectional symmetry in visual stimuli.

\begin{figure}[t]

\centering 
\noindent\begin{minipage}[b]{0.85\linewidth}
\centering
\scriptsize
\subcaptionbox*{\scriptsize\sffamily sym.: 0.87\\var.: 0.0122}[.24\textwidth][c]{\includegraphics[width=.24\textwidth]{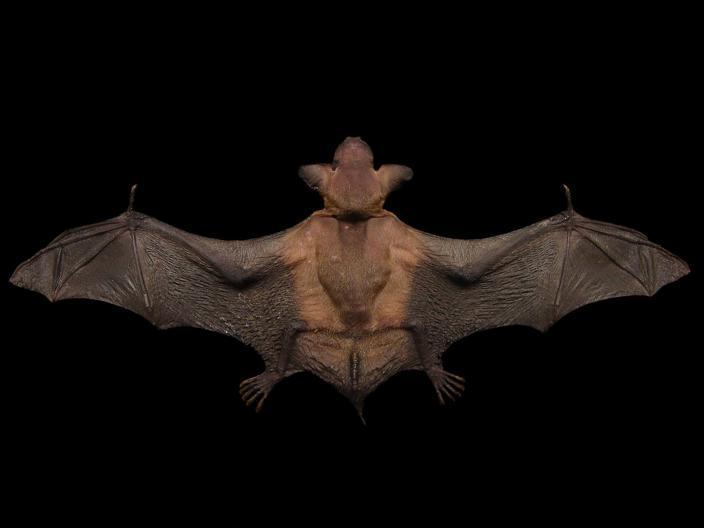}}
\subcaptionbox*{\scriptsize\sffamily sym.: 0.78\\var.: 0.1857}[.24\textwidth][c]{\includegraphics[width=.24\textwidth]{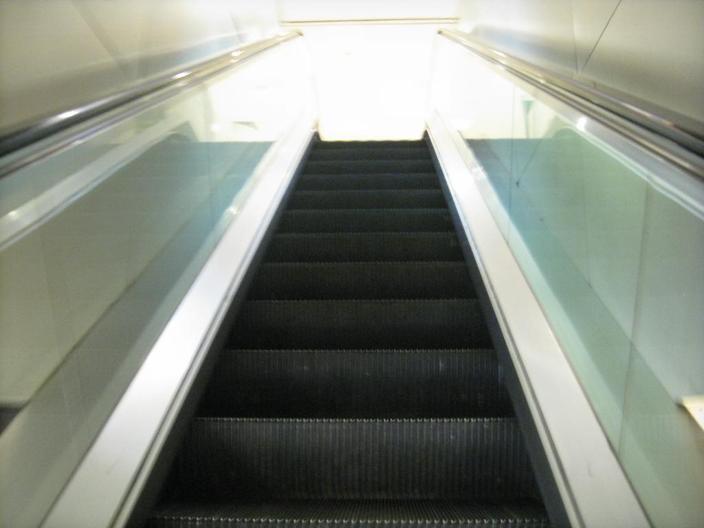}}
\subcaptionbox*{\scriptsize\sffamily sym.: 0.67\\var.: 0.1735}[.24\textwidth][c]{\includegraphics[width=.24\textwidth]{paper-assets/pics/QID122.jpg}}
\subcaptionbox*{\scriptsize\sffamily sym.: 0.64\\var.: 0.2224}[.24\textwidth][c]{\includegraphics[width=.24\textwidth]{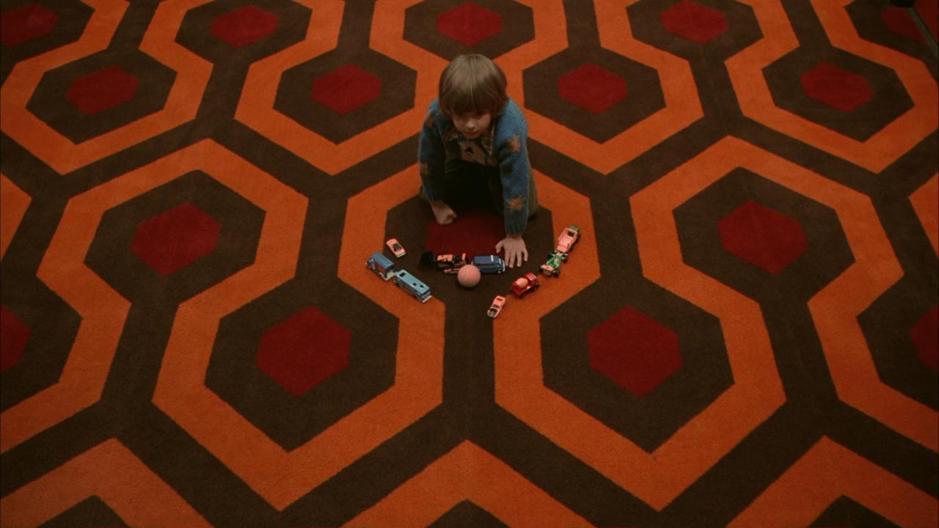}}

\smallskip

\subcaptionbox*{\scriptsize\sffamily sym.: 0.01\\var.: 0.0122}[.24\textwidth][c]{\includegraphics[width=.24\textwidth]{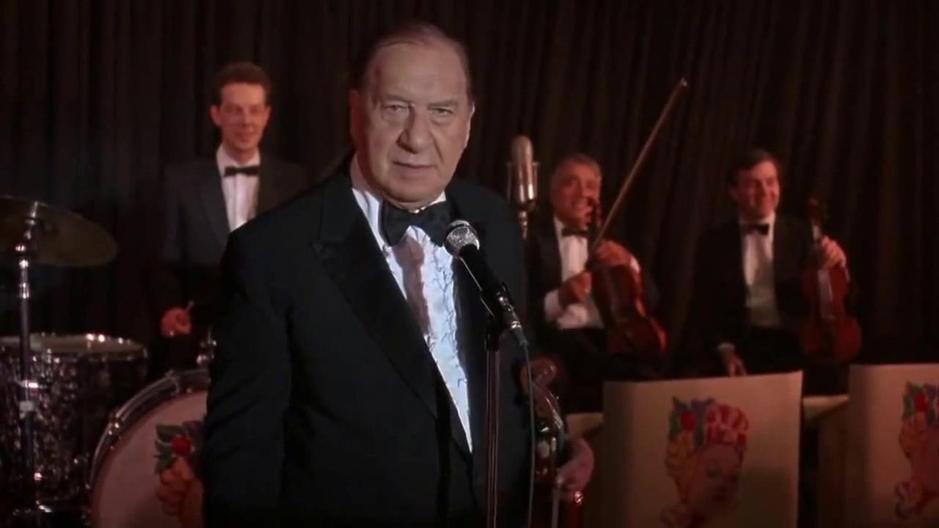}}
\subcaptionbox*{\scriptsize\sffamily sym.: 0.01\\var.: 0.0199}[.24\textwidth][c]{\includegraphics[width=.24\textwidth]{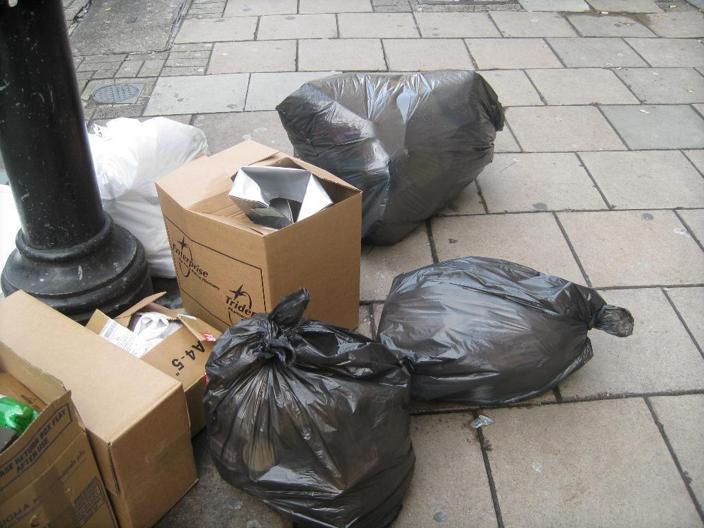}}
\subcaptionbox*{\scriptsize\sffamily sym.: 0.02\\var.: 0.0190}[.24\textwidth][c]{\includegraphics[width=.24\textwidth]{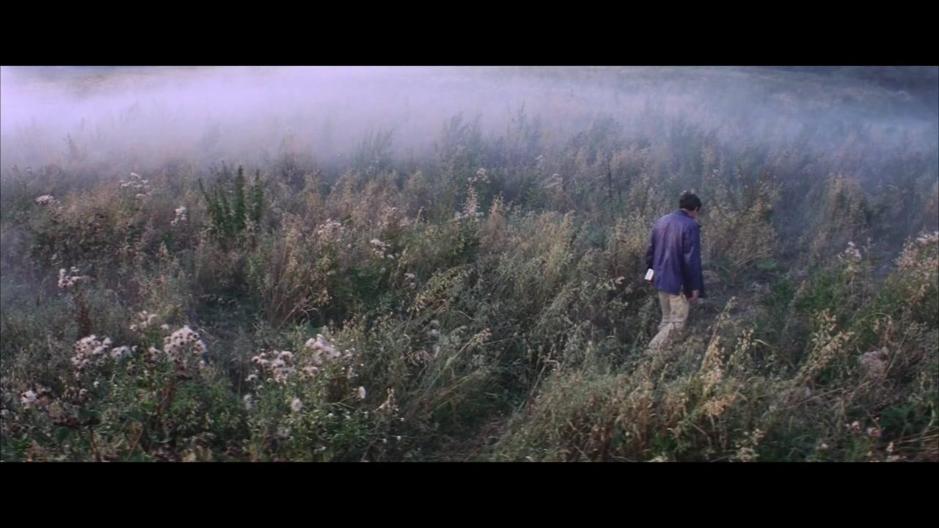}}
\subcaptionbox*{\scriptsize\sffamily sym.: 0.02\\var.: 0.0221}[.24\textwidth][c]{\includegraphics[width=.24\textwidth]{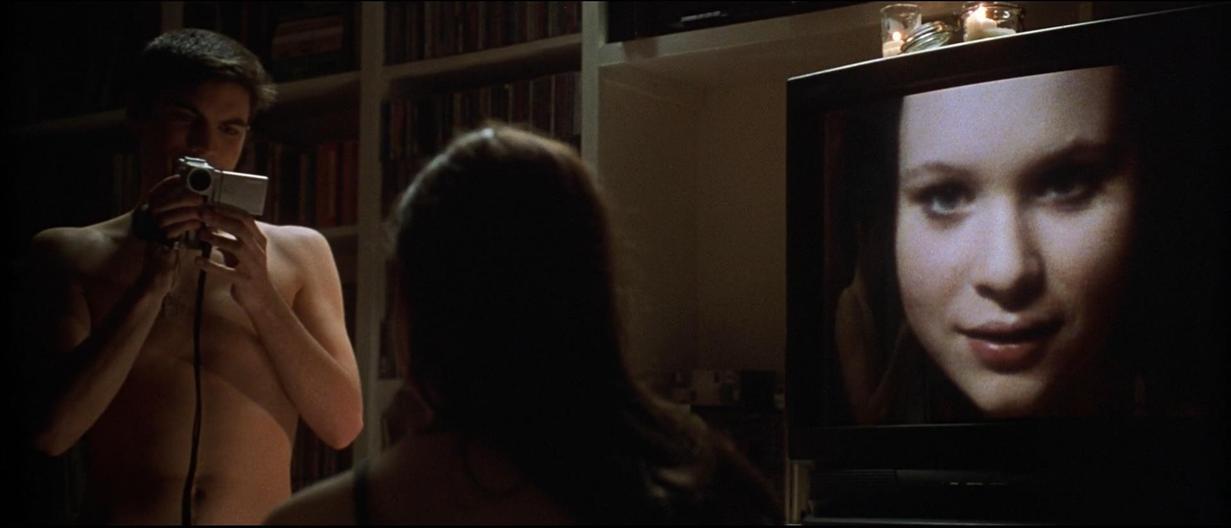}}

\smallskip

\subcaptionbox*{\scriptsize\sffamily sym.: 0.54\\var.: 0.4614}[.24\textwidth][c]{\includegraphics[width=.24\textwidth]{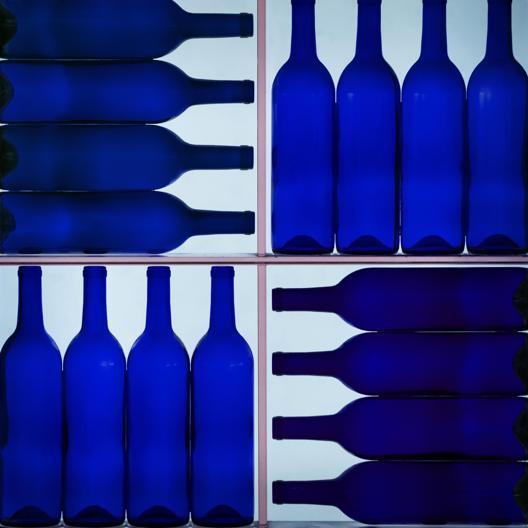}}
\subcaptionbox*{\scriptsize\sffamily sym.: 0.35\\var.: 0.4120}[.24\textwidth][c]{\includegraphics[width=.24\textwidth]{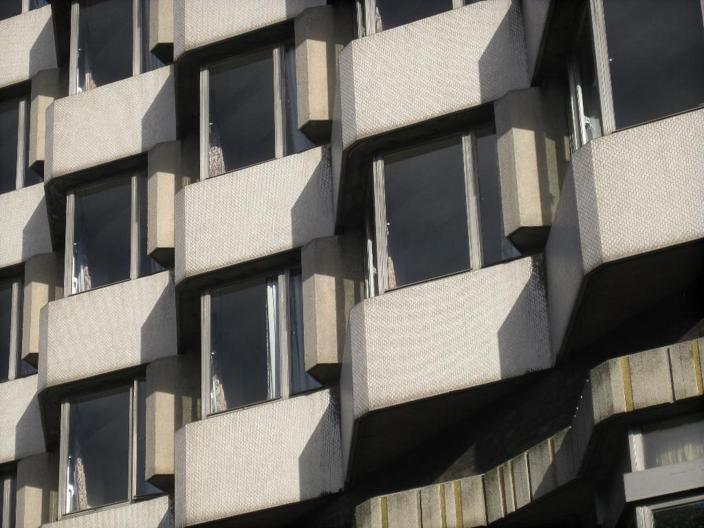}}
\subcaptionbox*{\scriptsize\sffamily sym.: 0.26\\var.: 0.3506}[.24\textwidth][c]{\includegraphics[width=.24\textwidth]{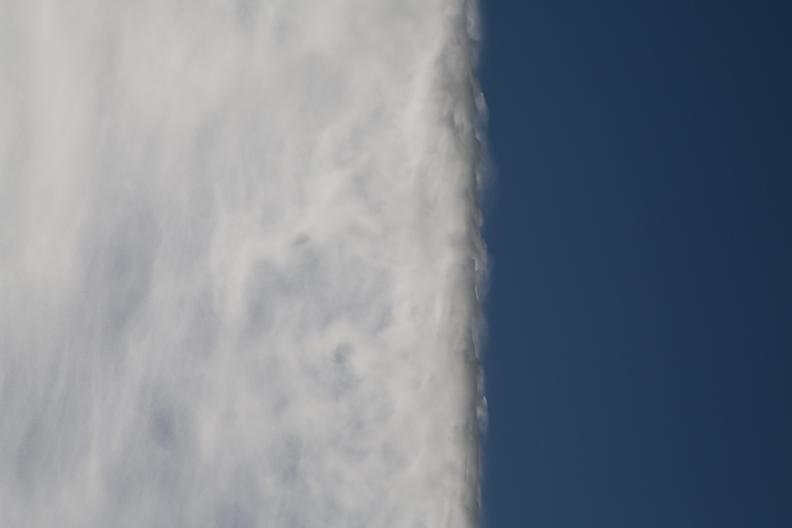}}
\subcaptionbox*{\scriptsize\sffamily sym.: 0.48\\var.: 0.3202}[.245\textwidth][c]{\includegraphics[width=.24\textwidth]{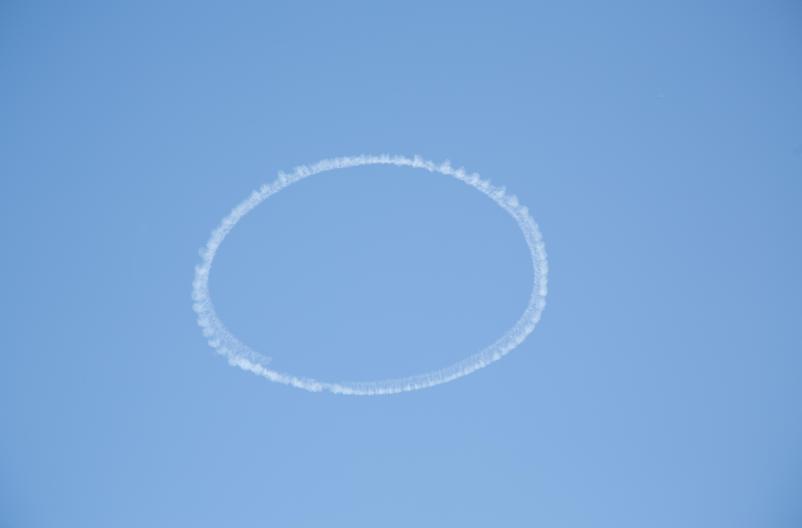}}

\end{minipage}

\caption{ Sample results from the human experiment.  ({\bf row 1}) most symmetric;\quad ({\bf row 2}) most non-symmetric (these correspond directly to the images with the lowest variance in the answers);\quad ({\bf row 3}) images with the biggest variance in the answers.}

\label{fig:human_study}

\end{figure}

\begin{figure}[t]
\begin{minipage}[b]{1.0\linewidth}
	\centering
    
    \subcaptionbox{\label{subfig:study_1}}{\includegraphics[height=2.15cm]{paper-assets/pics/QID130.jpg}}
    \subcaptionbox{\label{subfig:study_2}}{\includegraphics[height=2.15cm]{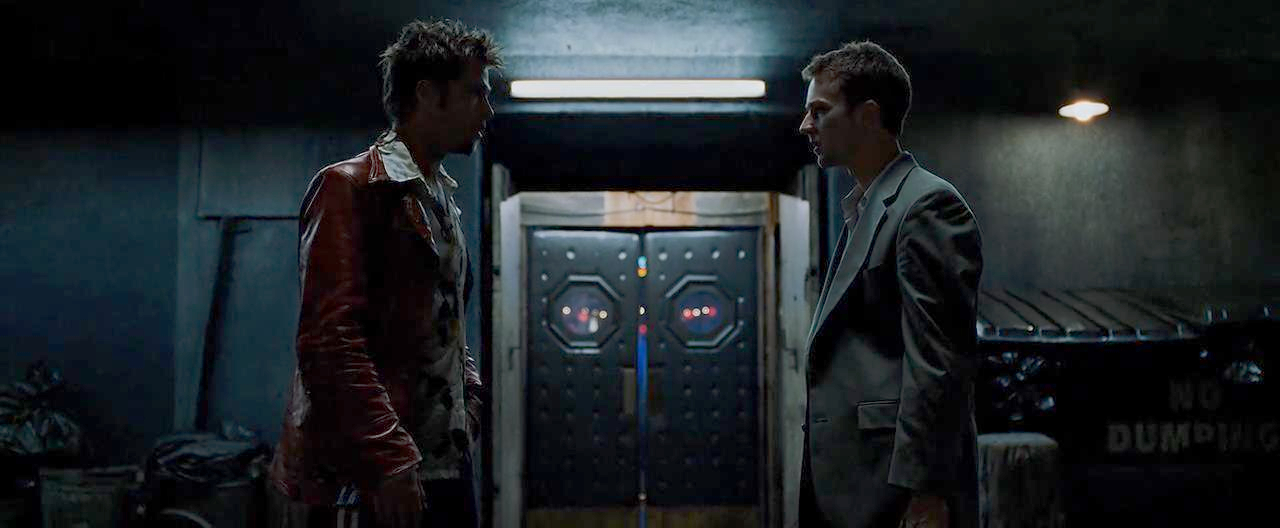}}
    \subcaptionbox{\label{subfig:study_3}}{\includegraphics[height=2.15cm]{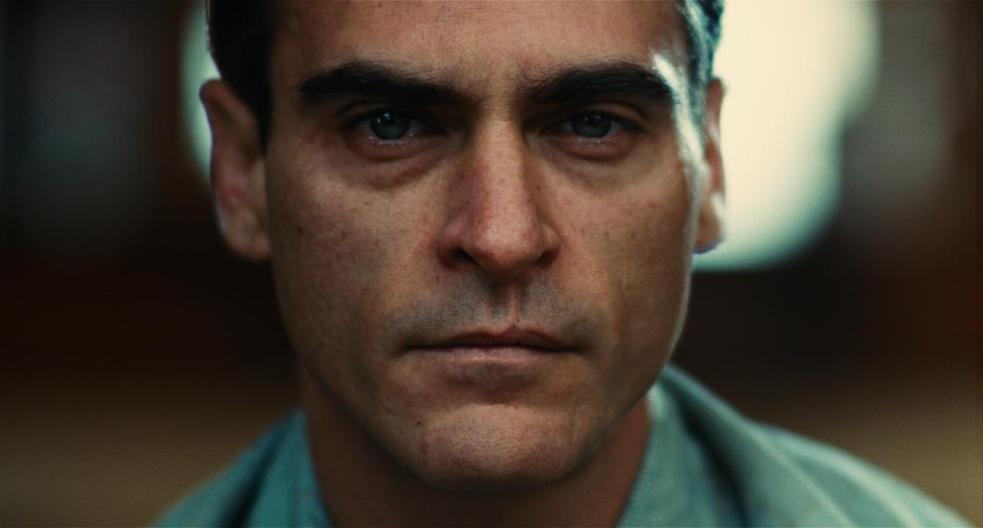}}

	\caption{{ Samples from the experimental data.}}

	\label{fig:qualitative_examples}
	
\end{minipage}
\end{figure}

\section{Human Evaluation: A Qualitative Study}\label{sec:human-eval}

{\bf Experimental Dataset} \quad Human-generated data from subjective, qualitative assessments of symmetry serves many useful purposes:  we built a dataset of $150$ images consisting of landscape and architectural photography, and movie scenes. The images range from highly symmetric images showing very controlled symmetric patterns to completely non symmetric images. Each participant was shown $50$ images selected randomly from the dataset; subjects had to rank the images by selecting one of four categories: {\footnotesize\sffamily not\_symmetric, somewhat\_symmetric, symmetric}, and {{\footnotesize\sffamily highly\_symmetric}. Each image was presented to approximately $100$ participants; we calculated the symmetry value as the average of all responses. 
 
\smallskip

{\bf Empirical Results} \quad The results from the human experiment suggest, that perception of symmetry varies a lot between subjects. While in the case of no symmetry people tend to agree, i.e. variance in the answers is very low (see Figure \ref{fig:human_study}), in the case of high symmetry, there is a wider variance in the human perception of symmetry. In particular in the case of images with an average level of symmetry the variance in the answers tends to be high. Qualitatively, there are various aspects on the subjective judgement of symmetry that we can observe in the human evaluation {{(1 -- 3)}}:\quad  (1) \emph{absence of features} decreases the subjective rating of symmetry, e.g., the image in Figure \ref{subfig:study_1} has a nearly perfect symmetry in the image features, but as there are only very few features that can be symmetrical people only perceived it as medium symmetrical, with a high variance in the answers;\quad (2) \emph{symmetrical placement of people} in the image has a higher impact on the subjective judgement of symmetry then other objects, e.g., the image in Figure \ref{subfig:study_2} is judged as symmetrical based on the placement of the characters and the door in the middle, but the objects on the left and right side are not very symmetrical; \quad (3) images that are \emph{naturally structured} in a symmetrical way are judged less symmetrical then those arranged in a symmetrical way, e.g., images of centred faces as depicted in Figure \ref{subfig:study_3}, are rated less symmetrical then other images with similar symmetry on the feature level.  

\smallskip

\begin{figure}[t]
	\centering

        \subcaptionbox{\label{subfig:graph_1}}{\includegraphics[height=3.2cm]{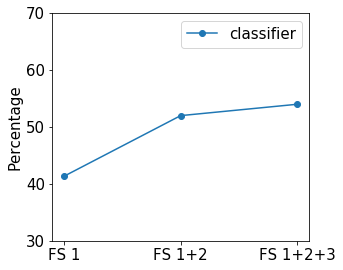}}\hfill
        \subcaptionbox{\label{subfig:graph_2}}{\includegraphics[height=3.2cm]{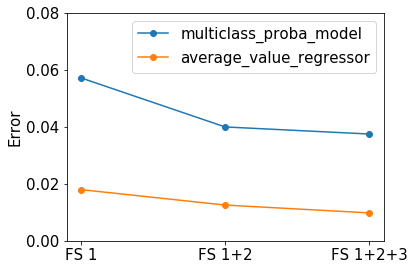}}\hfill
        \subcaptionbox{\label{subfig:graph_3}}{\includegraphics[height=3.2cm]{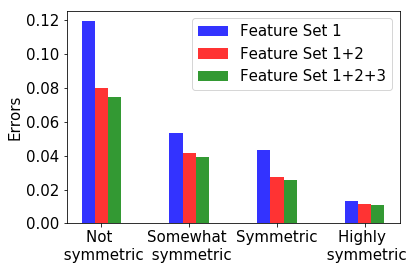}}

\caption{ Results of empirical evaluation with three different feature set combinations, showing (a)  mean accuracy, (b) mean error, and (c) class probability error. } 	
			
	\label{fig:Hist_All}
\end{figure}

\begin{table}[t]
\centering
\scriptsize
\begin{tabular}{|l|l|l|l|}
\hline
{\tiny\sffamily{\textbf{Feature Sets}}}					& {\tiny\sffamily{\textbf{CA (\%)}}}	&  {\tiny\sffamily{\textbf{Avg. Sym. Err.}}} &  {\tiny\sffamily{\textbf{Class Prob. Err.}}} \\ \hline
$fs1$            			& 41.33         		 & 0.01806876383            & 0.0572886659 \\ 
$fs1$+2     		& 52.00                      & 0.0126452444              & 0.0400713172 \\ 
$fs1$+2+3 	& 54.00                      & 0.009900461023          & 0.0375853705 \\ \hline
\end{tabular}
\caption{ Results from classification and prediction pipeline.}
\label{tbl:results}
\end{table}


{\bf Subjective symmetry interpretation}. \quad To evaluate how good our symmetry model reflects subjective human criteria for judging symmetry in naturalistic images, we use the results from the human study to train a classifier and a regressor to predict the symmetry class of an image and predict the average symmetry of the images.  For our experiment, we extracted three sets of features ($fs1$ - $fs3$) from the symmetry model:  $fs1$ consists of the cosine similarity between the two halves of each image on each of the $5$ convolution layers in an AlexNet;  $fs2$ consists of the symmetrical properties between image patches, i.e., divergence from symmetrical spatial configuration, and perceptual similarity; and $fs3$ consists of the symmetrical properties of object configuration and people in the images. We have 2 models, a classifier and a regressor. A given image is classified into one of the 4 symmetry classes using the classifier. This model is evaluated using the mean accuracy as shown in Figure \ref{fig:Hist_All}(a). The classifier model also predicts the per class probabilities, this is denoted by $multiclass\_proba\_model$. This model is evaluated by calculating the Mean Squared Error (MSE) between the predicted probabilities and the percentages from the human data for each class. The  per class errors are shown in Figure \ref{fig:Hist_All}(c) while the mean error is shown in Figure \ref{fig:Hist_All}(b). The regressor model predicts the average symmetry value of a given image. The model is evaluated by calculating the MSE between the predicted average symmetry value and average symmetry value from the human data. We use the pipeline optimization method of TPOT \citep{TPOT2016} to automatically build the classification and regression pipelines for the feature sets. This results in a classification pipeline consisting of an ensemble of DecisionTrees, SVM, RandomForest classifiers while the regression pipeline consists of an ensemble of ExtraTrees and XGBoost regressors. The models are trained and tested on the 3-feature set using 5-fold cross validation, splitting the 150 images into 5 folds. Reported are \emph{mean error} and \emph{classification accuracy} (CA).

\paragraph{Results and Discussion}

%
The results (Figure \ref{fig:Hist_All}; Table \ref{tbl:results}) show that using the features from our symmetry model improves performance in both tasks, i.e., the accuracy for the classification task improves by over 10\% (see Table \ref{tbl:results}) from 41.33 \% to 54\%, and for the per class probabilities the errors decreases from 0.057 to 0.038. 
The biggest improvement in the classification and in the prediction of the average symmetry value happens when adding the image patch features $fs2$ (Figure \ref{fig:Hist_All}(a), and (b)).  Adding people centred features only results in a small improvement, which may be because only a subset of the images in the dataset involves people. The results on the predicted per class probabilities (Figure \ref{fig:Hist_All}(c)) show that by adding features from our symmetry model we are able to make better predictions on the variances in the human answers. 





\section{Discussion and Related Work}
Symmetry in images has been studied from different perspectives, including visual perception research, neuroscience, cognitive science, arts and aesthetics \citep{Treder2010}. The semantic interpretation of symmetry from the viewpoint of perception and aesthetics requires a mixed empirical-analytical methodology consisting of both empirical and analytical methods: 

\begin{itemize}
	\item \textbf{ Empirical} / \emph{Human Behaviour Studies}.\quad This involves qualitative studies involving subjective assessments, as well as an evidence-based approach measuring human performance from the viewpwoint of visual perception using eye-tracking, qualitative evaluations, and think-aloud analysis with human subjects; and

	\item \textbf{ Analytical} / \emph{Interpretation and Saliency}.\quad This involves the development of computational models that serve an interpretation and a predictive function involving, for instance: (i)  multi-level computational modelling of interpreting visuo-spatial symmetry; \quad (ii) a saliency model of visual attention serving a predictive purpose vis-a-vis the visuo-spatial structure of visual media.

\end{itemize}



\paragraph{Symmetry and (computer) vision}

Symmetry is an important feature in visual perception and there are numerous studies in vision research investigating how symmetry affects visual perception \citep{Cohen2013,Norcia2002,Machilsen2009,Bertamini2014}, and how it is detected by humans \citep{Wagemans1997,Freyd1984,force-of-symm}. Most relevant to our work is the research on computational symmetry in the area of computer vision \citep{Liu_2013_CVPR_Workshops,CGV-008}. Typically, computational studies on symmetry in images characterise symmetry as \emph{reflection, translation, and rotation symmetry}; here,  reflection symmetry (also referred to as \emph{bilateral} or \emph{mirror symmetry}) has been investigated most extensively. Another direction of research in this area focuses on detecting symmetric structures in objects. In this context \cite{Teo2015} presents a classifier that detects curved symmetries in 2D images.  Similarly, \cite{Lee2012} presented an approach to detect curved glide-reflection symmetry in 2D and 3D images, and \cite{Atadjanov2016} uses appearance of structure  features to detect symmetric structures of objects. 

\paragraph{Computational analysis of image structure}
Analysing image structure is a central topic in computer vision research and there are various approaches for different aspects involved in this task. Deep learning with convolutional neural networks (CNNs) provide the basis for analysing images using learned features, e.g., AlexNets \citep{krizhevsky2012imagenet}, or ResNets\citep{He2016_resnet}, trained on the ImageNet Dataset \citep{imagenet2009}.
 Most recent developments in object detection involve \emph{RCNN} based detectors such as \cite{Ren17} and \cite{Girshick16}, where objects are detected based on region proposals extracted from the image, e.g., using selective search \citep{Uijlings2013} or region proposal networks for predicting object regions. For comparing images, \cite{Zagoruyko2015} and \cite{DB16c} measure perceptual similarity based on features learned by a neural network.

\begin{figure}[t]

\centering

\scriptsize

\begin{subfigure}[b]{\linewidth}
	\centering
    
    \includegraphics[width=\columnwidth]{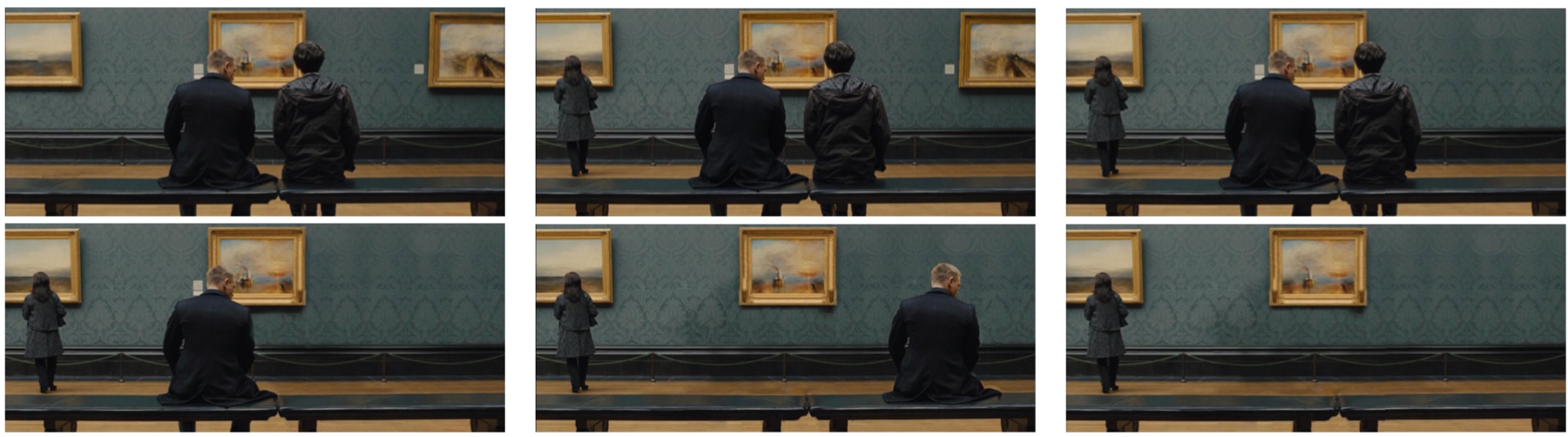}\quad

\end{subfigure}

\caption{{ Manual manipulation of symmetry. Symmetry decreasing from  highly symmetric to not symmetric.}}

	\label{fig:symmetry_manipulation}
\end{figure}%

\section{Summary and Outlook}\label{sec:outlook}
Our research addresses visuo-spatial symmetry in the context of naturalistic stimuli in the domain of visual arts, e.g., {film, paintings, and landscape and architectural photography}. With a principal focus on developing a human-centred computational model of (interpreting) visuo-spatial symmetry, our approach is  motivated and driven by three crucial and mutually synergistic aspects, namely: reception, {interpretation}, and {synthesis}:

\begin{itemize}
\item {\bf Reception}:\quad A behavioural study of the human perception (and explanation) of symmetry from the viewpoint of visual attention, and spatio-linguistic and qualitative characterisation(s);

\item {\bf Interpretation}:\quad A computational model of deep semantic interpretation of visual symmetry with an emphasis on human-centred explainability and visual sensemaking;

\item {\bf Synthesis}:\quad The ability to apply human-centred explainable models as a basis to directly or indirectly engineer visual media vis-a-via their (predictive) receptive effects, i.e., guiding attention by influencing visual fixation patterns, minimising / maximising saccadic movements (e.g., in animation, gaming, built environment planning, and design). 
\end{itemize}

In this paper, we have focussed on the reception and interpretation aspects; we presented a declarative, computational model of reflectional symmetry integrating (visuospatial) composition, feature-level similarity, and semantic similarity in visual stimuli. Some possible next steps could be:

\begin{itemize}

\item  \textbf{spatio-temporal symmetry and visual perception}: Going beyond static images to analyse symmetry in \emph{space-time} (e.g., as in the films of {\sffamily\small Wes Anderson} \citep{Bhatt-Suchan-Symmetry-SHAPES2015}): here, a particular focus is on the influence of space-time symmetry on visual fixations and saccadic eye-movements \citep{DBLP:conf/apgv/SuchanBY16}

\item  \textbf{visual processing aspect}: More advanced region proposals are possible, and can be naturally driven by newer forms of visual computing primitives and similarity measures. The framework is modular and may be extended with improved or new visual-computing features
 
\item \textbf{Resynthesising images}: produce qualitatively distinct classes of (a)symmetry (e.g., Figure \ref{fig:symmetry_manipulation}), and conducting further empirical studies involving qualitative surveys, eye-tracking, think-aloud studies etc

\end{itemize}

The most immediate outlook of our research on the computational front is geared towards extending the current symmetry model for the analysis of \textbf{spatio-temporal symmetry} particularly from the viewpoint of moving images as applicable in film, animation, and other kinds of narrative media. Towards this, we extend the symmetry model to include a richer spatio-temporal ontology, e.g., consisting of `\emph{space-time}' entities \citep{DBLP:conf/ijcai/SuchanB16,DBLP:conf/aaai/SuchanBWS18,ASP-MT-Space-Time-RuleML2018} for the analysis of spatio-temporal symmetry. Space-time symmetry analysis will also be complemented with specialised methods that provide a holistic view of the cinematographic  ``\textbf{geometry of a scene}'' \citep{DBLP:conf/wacv/SuchanB16,DBLP:conf/ijcai/SuchanB16}. Another promising line of work we are exploring involves relational learning of visuo-spatial symmetry patterns (e.g., based on based on inductive generalisation \citep{DBLP:conf/ilp/SuchanBS16}). Explainable learning from (big) visual datasets promises to offer a completely new approach towards the study of media and art history, cultural studies, and aesthetics.

\section*{Acknowledgements}
We thank the anonymous reviewers and ACS 2018 program chairs Pat Langley and Dongkyu Choi for their constructive comments and for helping improve the presentation of the paper. The empirical aspect of this research was made possible by the generous participation of of volunteers in the online survey; we are grateful to all participants. We acknowledge the support of Steven~Kowalzik towards digital media work pertaining to manual re-synthesis of categories of stimuli.


\bibliographystyle{cogsysapa}

\end{document}